\documentclass{article}

\usepackage[preprint]{neurips_2025}

\usepackage[utf8]{inputenc} % allow utf-8 input
\usepackage[T1]{fontenc}    % use 8-bit T1 fonts
\usepackage{hyperref}       % hyperlinks
\usepackage{url}            % simple URL typesetting
\usepackage{booktabs}       % professional-quality tables
\usepackage{amsfonts}       % blackboard math symbols
\usepackage{nicefrac}       % compact symbols for 1/2, etc.
\usepackage{microtype}      % microtypography
\usepackage{xcolor}         % colors

\usepackage{amsmath}
\usepackage{amssymb}
\usepackage{mathtools}
\usepackage{amsthm}

\theoremstyle{plain}

\usepackage{amsfonts} 
\usepackage{xcolor}
\usepackage{subfigure}
\usepackage{caption}
\usepackage{float}
\usepackage{wrapfig}
\usepackage{graphicx}
\usepackage{subcaption}

\usepackage{multirow}
\usepackage{array}
\usepackage{tabularx}
\usepackage{amssymb}
\usepackage{booktabs}
\usepackage{xcolor}
\usepackage{graphicx}
\usepackage{array}

\usepackage{amsmath, amsthm}
\usepackage{algorithm}
\usepackage{algorithmic}
\usepackage{pythonhighlight}

\newtheorem{definition}{Definition}

\newcommand{\cmark}{\textcolor{green}{\scalebox{1.2}{\textbf{\checkmark}}}}
\newcommand{\xmark}{\textcolor{red}{\scalebox{1.2}{\textbf{$\times$}}}}

\def\HiLi{\leavevmode\rlap{\hbox to \hsize{\color{yellow!50}\leaders\hrule height .8\baselineskip depth .5ex\hfill}}}

\title{FlashDP: Private Training Large Language Models with Efficient DP-SGD}

\author{%
  Liangyu Wang\\
  King Abdullah University of Science and Technology\\
  \texttt{liangyu.wang@kaust.edu.sa}\\
  \And
  Junxiao Wang\\
  Guangzhou University\\
  \texttt{junxiao.wang@gzhu.edu.cn}\\
  \And
  Jie Ren\\
  King Abdullah University of Science and Technology\\
  \texttt{jie.ren@kaust.edu.sa}\\
  \And
  Zihang Xiang\\
  King Abdullah University of Science and Technology\\
  \texttt{zihang.xiang@kaust.edu.sa}\\
  \And
  David E. Keyes\\
  King Abdullah University of Science and Technology\\
  \texttt{david.keyes@kaust.edu.sa}\\
  \And
  Di Wang\\
  King Abdullah University of Science and Technology\\
  \texttt{di.wang@kaust.edu.sa}\\
}

\begin{document}

\maketitle

\begin{abstract}
As large language models (LLMs) increasingly underpin technological advancements, the privacy of their training data emerges as a critical concern. Differential Privacy (DP) serves as a rigorous mechanism to protect this data, yet its integration via Differentially Private Stochastic Gradient Descent (DP-SGD) introduces substantial challenges, primarily due to the complexities of per-sample gradient clipping. Current explicit methods, such as Opacus, necessitate extensive storage for per-sample gradients, significantly inflating memory requirements. Conversely, implicit methods like GhostClip reduce storage needs by recalculating gradients multiple times, which leads to inefficiencies due to redundant computations. This paper introduces FlashDP, an innovative cache-friendly per-layer DP-SGD that consolidates necessary operations into a single task, calculating gradients only once in a fused manner. This approach not only diminishes memory movement by up to \textbf{50\%} but also cuts down redundant computations by \textbf{20\%}, compared to previous methods. Consequently, FlashDP does not increase memory demands and achieves a \textbf{90\%} throughput compared to the Non-DP method on a four-A100 system during the pre-training of the Llama-13B model, while maintaining parity with standard per-layer clipped DP-SGD in terms of accuracy. These advancements establish FlashDP as a pivotal development for efficient and privacy-preserving training of LLMs. 
FlashDP's code has been open-sourced in \href{https://github.com/kaustpradalab/flashdp}{\textit{https://github.com/kaustpradalab/flashdp}}.
\end{abstract}

\section{Introduction}

The transformer architecture \citep{vaswani2017attention} has revolutionized fields like natural language processing \citep{gao2024llm,xie2023translating}, embodied AI \citep{song2023llm,duan2022survey,xu2024survey}, and AI-generated content (AIGC) \citep{cao2023comprehensive,wu2023ai}, with Large Language Models (LLMs) demonstrating exceptional abilities in text generation, complex query responses, and various language tasks due to training on massive datasets. These models, exemplified by ChatGPT, are applied across diverse areas, including healthcare, where they enhance diagnosis and drug discovery by analyzing medical data \citep{toma2023clinical,ali2023using,sheikhalishahi2019natural,sallam2023chatgpt,biswas2023role}. However, the extensive capabilities of LLMs raise significant privacy concerns, particularly as they can inadvertently expose or generate sensitive information, owing to their potential to memorize data from large training sets \citep{pang2024bolt,nasr2023scalable,carlini2023quantifying,ippolito2022preventing,mccoy2023much,tirumala2022memorization,zhang2023counterfactual,ashkboos2023towards}.

Differential Privacy (DP) ensures privacy by adding noise during data processing, such that any single data point's influence on outcomes is minimal \citep{dwork2006differential}. As the most commonly adopted methods for ensuring  DP in deep learning models, Differentially Private Stochastic Gradient Descent (DP-SGD)  based methods \citep{abadi2016deep} adapt traditional stochastic gradient descent by clipping gradients per sample and adding noise. Although DP-SGD's application in LLMs is increasing, recent research \citep{li2022large,bu2023differentially,anil2022large,hoory2021learning} primarily targets the fine-tuning phase, providing privacy only for fine-tuned data. While some studies \citep{lee2021scaling,li2022large,bu2023differentially} have applied DP-SGD to pre-training, they often exhibit limited scalability or reduced training efficiency. This is primarily due to the significant computational and memory overheads inherent to per-sample gradient processing in DP-SGD, which make end-to-end pre-training of large models particularly challenging.

\begin{wrapfigure}{r}{0.6\textwidth}
  \begin{center}
    \includegraphics[width=0.58\textwidth]{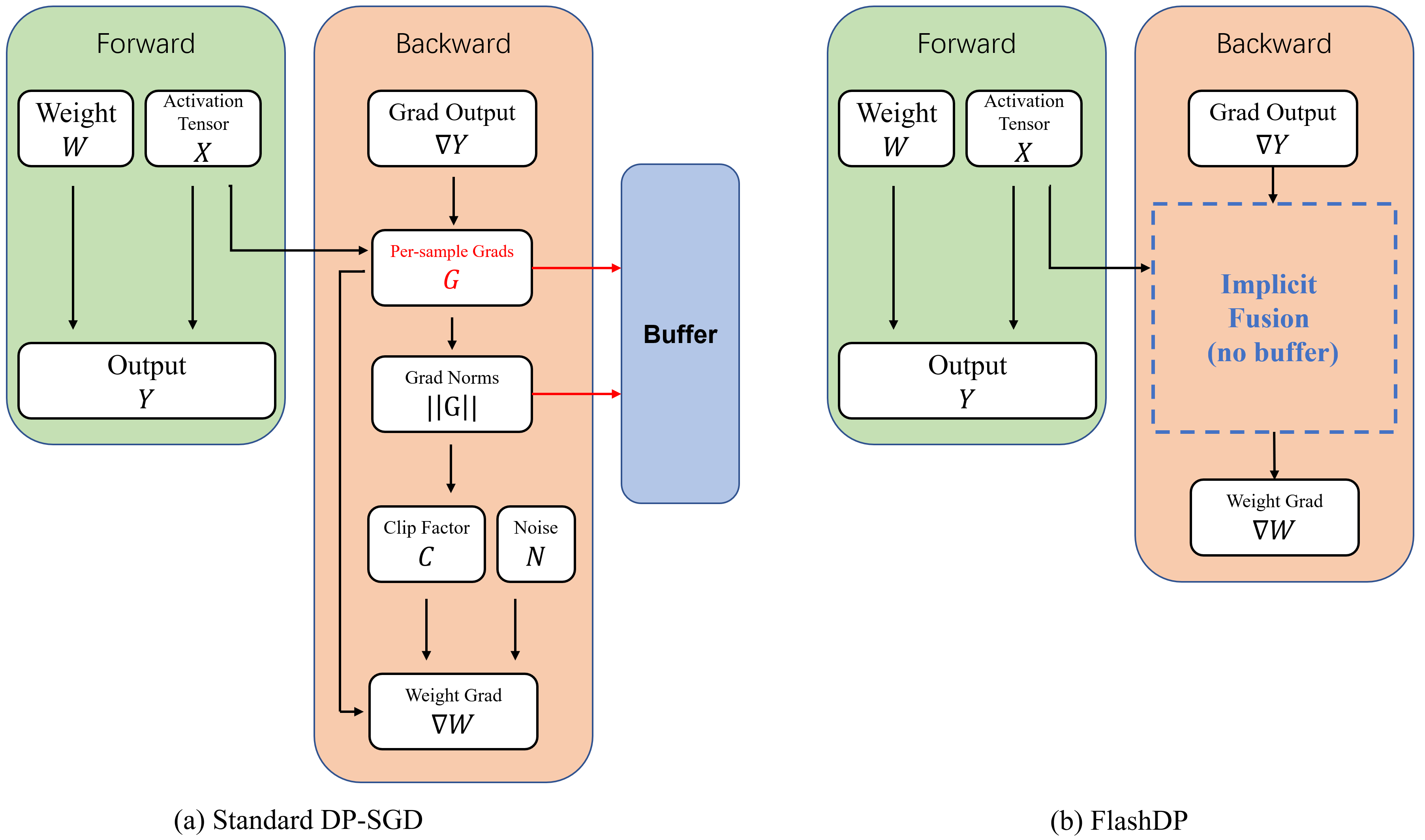}
  \end{center}
  \caption{Comparison of different training methods. (a) Standard DP-SGD: Stores per-sample gradients $\mathbf{G}$ (red explicit cache), increasing memory usage (blue buffer). (b) FlashDP: Optimizes gradient processing by consolidating computations into a single pass, reducing redundancy and memory use.}
    \label{fig:intro}
\end{wrapfigure}

Integrating DP into LLM training via DP-SGD/Adam poses significant challenges, particularly due to per-sample gradient clipping. This crucial privacy technique involves adjusting each data sample's gradients to limit their influence on model updates. While critical for maintaining strict privacy standards, this approach requires computing and storing individual gradients, significantly raising computational and memory demands. Managing these gradients is especially taxing in LLMs, which are known for their large parameter spaces. Each gradient must be carefully clipped and aggregated before updating model parameters, straining computational resources, and prolonging training times. These scalability issues are particularly acute in settings with limited hardware, creating significant barriers to efficiently training privacy-aware LLMs \citep{li2022large,bu2023differentially}.

Current research on DP-SGD for training LLMs can be categorized into two classes:  explicit methods like Opacus \citep{yousefpour2021opacus} stand out by directly storing per-sample gradients. This approach, while straightforward, significantly increases the memory footprint (Appendix Table \ref{tab:comparison}), which becomes prohibitive for state-of-the-art LLMs characterized by billions of parameters \citep{touvron2023llama, achiam2023gpt}. Such a substantial increase in memory requirements hampers scalability and renders these methods impractical for deployment in large-scale model training environments. The direct storage of gradients, essential for ensuring the privacy guarantees of DP, thus poses a substantial barrier to the efficient implementation of DP in LLMs.

Conversely, implicit methods, exemplified by innovations such as GhostClip \citep{li2021large}, address the memory challenge by circumventing the need for persistent storage of per-sample gradients. These methods segment the DP-SGD process into multiple discrete computational tasks, ostensibly to mitigate memory demands. However, this strategy necessitates the frequent recalculation of per-sample gradients, which introduces a high degree of computational redundancy (Table \ref{tab:comparison}). This redundancy not only undermines training efficiency but also extends the duration of the training process significantly. For  LLMs, which require substantial computational resources and extended training times, the inefficiencies introduced by such redundant computations become a critical bottleneck. These implicit methods, while innovative in reducing memory usage, thus struggle to deliver a practical solution for the privacy-preserving training of LLMs at scale.

To effectively tackle the challenges presented by existing methods of integrating DP into the training of LLMs, we introduce FlashDP, a novel, cache-friendly implicit algorithm designed to streamline the per-layer clipping DP-SGD process (Figure \ref{fig:intro} (a)). 
We opt for per-layer clipping in our research primarily due to its efficiency in managing both memory consumption and accuracy, especially vital in the differentially private training of expansive language models \citep{bu2023accuracy,he2022exploring}. It has been shown that this type of method not only sustains commendable accuracy compared to the standard DP-SGD but also mitigates memory overhead, a critical consideration when training large-scale models under privacy constraints. 
FlashDP uniquely implements a unified computational strategy that performs the gradient operations required for DP-SGD in a single pass (Figure \ref{fig:intro} (b)). This innovative approach not only eliminates the need for multiple recalculations of per-sample gradients but also consolidates the entire process into one cohesive computational task. To be specific, FlashDP's architecture, which consolidates the entire DP-SGD process into a single GPU kernel, eliminates redundant computations and optimizes data flow within the GPU. This integration results in a streamlined workflow that efficiently manages memory and processing resources. Also, FlashDP reorganizes the GPU operations to maximize data throughput and minimize latency, effectively enhancing the overall efficiency of the training process. These architectural improvements significantly reduce the volume of memory transfers and computational redundancies, thereby optimizing both the speed and resource utilization during the training of LLMs with DP.

By re-designing the gradient computation workflow, FlashDP dramatically reduces the volume of memory transfers by 50\% and decreases redundant computational tasks by 20\% compared to previous implicit methods. This optimization is achieved through an advanced caching mechanism that efficiently manages gradient data and computation within GPU memory, minimizing the data movement across the system. As a result, FlashDP significantly alleviates the memory overhead traditionally associated with DP-SGD, enhancing the model's scalability and training speed.

The practical impact of these improvements is substantial. On a computational platform equipped with four NVIDIA A100 GPUs, FlashDP achieves a remarkable 90\% throughput compared to the non-DP method during the pre-training phase of the Llama-13B model, a state-of-the-art LLM known for its extensive data and computation demands. Crucially, this enhanced performance is attained without any degradation in the accuracy or dilution of the privacy guarantees compared to the original per-layer clipped DP-SGD. FlashDP thus not only meets but exceeds the operational requirements for effective and efficient privacy-preserving training of LLMs.

Our contributions can be summarized as follows:

\begin{itemize}
    \item {\bf Enhanced Throughput for LLM training with DP}: We propose FlashDP, which effectively resolves the issue of low throughput in DP-SGD/Adam with per-layer clipping during the training of LLMs. By optimizing the computational workflow and integrating more efficient handling of per-sample gradients, FlashDP significantly enhances the processing speed without compromising the model's accuracy or privacy integrity.
    \item {\bf Innovative GPU I/O Optimization}: Our study pioneers the exploration of DP-SGD from the perspective of GPU input/output operations. FlashDP's architecture, which consolidates the entire DP-SGD process into a single GPU kernel, eliminates redundant computations and optimizes data flow within the GPU. This approach not only reduces the computational load but also minimizes the number of GPU memory accesses, setting a new standard for efficiency in DP implementations.
    \item {\bf Experimental Validation of Efficiency and Scalability}: In practical LLM models involving \underline{Llama-13B}, FlashDP matches the speed and memory usage of non-DP training methods and achieves a significant {\bf 90\%} throughput compared with Non-DP methods. This performance is achieved on a computational platform equipped with four NVIDIA A100 GPUs. Importantly, it accomplishes this without any degradation in the precision or the privacy guarantees typically observed in the previous per-layer clipped DP-SGD implementations. This capability demonstrates FlashDP's effectiveness in scaling DP applications to larger and more complex LLMs without the usual trade-offs.
\end{itemize}

\section{Related Work}

\noindent {\bf Improving Time and Memory Complexities of DP-SGD.} The transition from standard stochastic gradient descent to DP-SGD introduces substantial modifications in memory and computational demands. In conventional settings, parameter updates are efficiently computed by aggregating gradients across all samples within a batch. This approach is both memory-efficient and computationally straightforward. In contrast, DP-SGD mandates that each sample's gradients be preserved, clipped, and subsequently aggregated to uphold privacy guarantees. 
Recent innovations in DP-SGD have primarily concentrated on ameliorating its computational and memory inefficiencies.
TF-Privacy vectorizes the loss to calculate per-sample gradients through backpropagation, which is efficient in terms of memory but slow in execution \citep{tensorflow2015-whitepaper}. Opacus \citep{yousefpour2021opacus} and \citep{rochette2019efficient} enhance the training efficiency by employing the outer product method \citep{goodfellow2015efficient}, albeit at the cost of increased memory usage needed to store per-sample gradients. This memory overhead is mitigated in FastGradClip \citep{lee2020scaling} by distributing the space complexity across two stages of backpropagation, effectively doubling the time complexity. Additionally, ghost clipping techniques \citep{goodfellow2015efficient}, \citep{li2021large}, \citep{bu2022scalable} allow for clipping per-sample gradients without full instantiation, optimizing both time and space, particularly when feature dimensions are constrained. Furthermore, \citep{bu2023differentially} introduces a 'book-keeping' (BK) method that achieves high throughput and memory efficiency, but still leaves room for improvement in fully addressing the computational and memory bottlenecks inherent in large-scale DP training.

While these methodologies have made significant strides in mitigating the extensive computational and memory demands typically associated with managing per-sample gradients in DP-SGD, they have not addressed the optimization of DP training from the perspective of GPU architecture and memory access. Additionally, the approaches detailed thus far do not cater effectively to the training of today's LLMs. FlashDP aims to enhance the efficiency and feasibility of training LLMs under the constraints of differential privacy, ensuring both high performance and adherence to privacy standards.

\noindent {\bf DP for Large Language Models.}  The field of privacy-preserving LLMs is characterized by the use or exclusion of DP and its extensions. 
\citep{he2022exploring} evaluated the precision equivalence of per-layer clipping with flat clipping on LLMs.
\citep{kerrigan2020differentially} demonstrated that public pretraining could facilitate downstream DP fine-tuning, although they did not explore fine-tuning large pre-trained models using DP-SGD. \citep{qu2021privacy} explored the fine-tuning of BERT for language understanding tasks under local DP. \citep{bommasani2021opportunities} suggested the potential for cost-effective private learning through fine-tuning large pre-trained language models. \citep{anil2021large} and \citep{dupuy2022efficient} extended these studies to BERT, pretraining and fine-tuning under global DP, respectively, with \citep{anil2021large} addressing datasets comprising hundreds of millions of examples, and \citep{dupuy2022efficient} reporting on datasets of utterances with relatively high $\epsilon$ values. 
Our research distinguishes itself by focusing on pre-training and fine-tuning large language models with high throughput and low memory usage.

\section{Understanding the Limitations of Previous Methods}

\begin{figure*}[!htbp]
    \centering
    \includegraphics[width=0.9\textwidth]{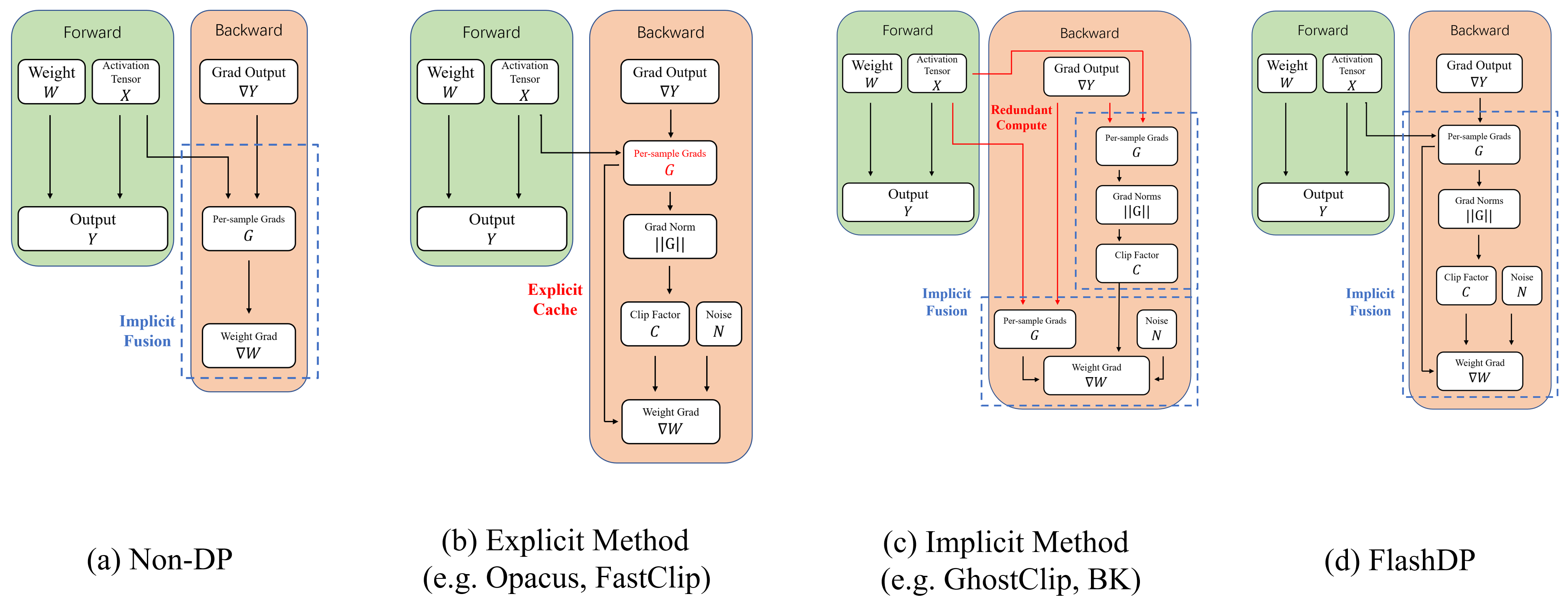}
    \caption{
    Comparison of different training methods. (a) Non-DP: Basic training without DP. (b) Explicit Method (e.g., Opacus, FastClip): Stores per-sample gradients $\mathbf{G}$ (red explicit cache), increasing memory usage. (c) Implicit Method (e.g., GhostClip, BK): Reduces memory by recalculating gradients in fused manners (blue dotted box) but implicitly calculating the per-sample gradient twice, causing computational redundancy. (d) FlashDP: Optimizes gradient processing by consolidating computations into a single pass, reducing redundancy and memory use. 
    }
    \label{fig:compare}
\end{figure*}

In this section, we introduce the previous non-DP, explicit, and implicate methods of DP-SGD from the GPU I/O perspective to
see their weakness, which motivates our framework. Due to the space limit, please refer to Appendix \ref{sec:pre} for the background on DP, Transformers, GPU architecture, and CUDA programming.  As discussed in Section \ref{sec:background-transformers}, the linear operation is crucial in the architecture of LLMs, particularly within Multi-Head Attention (MHA) and Feedforward Network (FFN) modules. Given its significance, we utilize the linear operation as an exemplar to elucidate the training workflow on GPUs, as shown in Figure \ref{fig:compare}. See Appendix \ref{sec:work_all} for details.

In the standard non-private training workflow of a linear layer, the forward pass involves a matrix multiplication \(Y = XW^{\mathsf{T}}\) between the activation tensor \(X \in \mathbb{R}^{B \times T \times P}\) and the weight matrix \(W \in \mathbb{R}^{D \times P}\), resulting in the output \(Y \in \mathbb{R}^{B \times T \times D}\), where \(B\), \(T\), \(P\), and \(D\) denote the batch size, sequence length, input feature dimension, and output feature dimension, respectively. The backward pass calculates the output gradient \(\nabla_Y \in \mathbb{R}^{B \times T \times D}\) and the weight gradient \(\nabla_W \in \mathbb{R}^{D \times P}\) via \(\nabla_W = \sum_{B} \sum_{T} (\nabla_Y)^{\mathsf{T}} X\). Figure~\ref{fig:compare} (a) illustrates this process, showing that the activation tensor \(X\) and weights \(W\) are stored in HBM for efficient access during computations, while intermediate operations utilize SRAM to enhance memory access time and throughput.

The explicit DP-SGD workflow, as depicted in Figure \ref{fig:compare} (b), categorizes the process into four stages to ensure privacy adherence by explicitly managing per-sample gradients. \textbf{Stage 1} involves computing per-sample gradients \(\mathbf{G} = \sum_T \nabla_Y^T X\) using batched GEMM operations on SRAM to minimize latency, with subsequent storage of the gradients back to HBM. \textbf{Stage 2} requires reloading these gradients to compute their norm \(\|\mathbf{G}\| = \sqrt{\sum_D \sum_P {\mathbf{G}^2}}\), then storing the results back in HBM. \textbf{Stage 3} includes loading the gradients and their norms for the per-layer clipping operations, ensuring that no gradient norm exceeds the predefined threshold \(C\), with the clipped gradients \(\mathbf{G}'\) written back to HBM. \textbf{Stage 4} focuses on adding Gaussian noise to the clipped gradients in SRAM for privacy preservation, followed by their aggregation for model updates, and storing the final noisy gradient \(\nabla_W\) back in HBM. This explicit handling of per-sample gradients not only increases memory usage but also complicates processing due to frequent memory swaps and disrupts efficient GPU utilization by breaking down kernel fusion strategies, becoming notably impractical for LLMs with their extensive parameter and gradient sizes, severely impacting training efficiency.

The implicit DP-SGD workflow, illustrated in Figure~\ref{fig:compare} (c), employs a method such as GhostClip to recalculate gradients in a fused manner, thus circumventing the need for explicit storage of per-sample gradients. \textbf{Stage 1} consolidates the first three stages of the explicit method into a single fused computational step, where the activation tensor \(X\) and output gradient tensor \(\nabla_Y\) are loaded into SRAM. Per-sample gradient tensor \(\mathbf{G}\) recalculations, norm calculations, and the per-layer clipping are integrated into one operation, minimizing latency and avoiding repeated data transfers to HBM. \textbf{Stage 2} mirrors the explicit method's final stage, where the recalculated and clipped gradients \(\mathbf{G}'\) undergo Gaussian noise addition in SRAM, followed by aggregation and storage in HBM for model updates. This approach reduces memory usage but increases computational load due to the redundancy of multiple gradient recalculations, which can significantly extend training times, rendering the method less practical due to the increased time complexity proportional to \(T\).

To address the previous limitations, the subsequent section will introduce FlashDP, a novel strategy designed to address these inefficiencies by rethinking the execution pipeline of DP-SGD. Without delving into specifics here, FlashDP's architecture will streamline the integration of per-sample gradient computation and clipping, potentially reducing the operational bottlenecks observed in existing methods.

\section{FlashDP Algorithm Design}

\begin{figure*}[!htbp]
    \centering
    \includegraphics[width=0.8\textwidth]{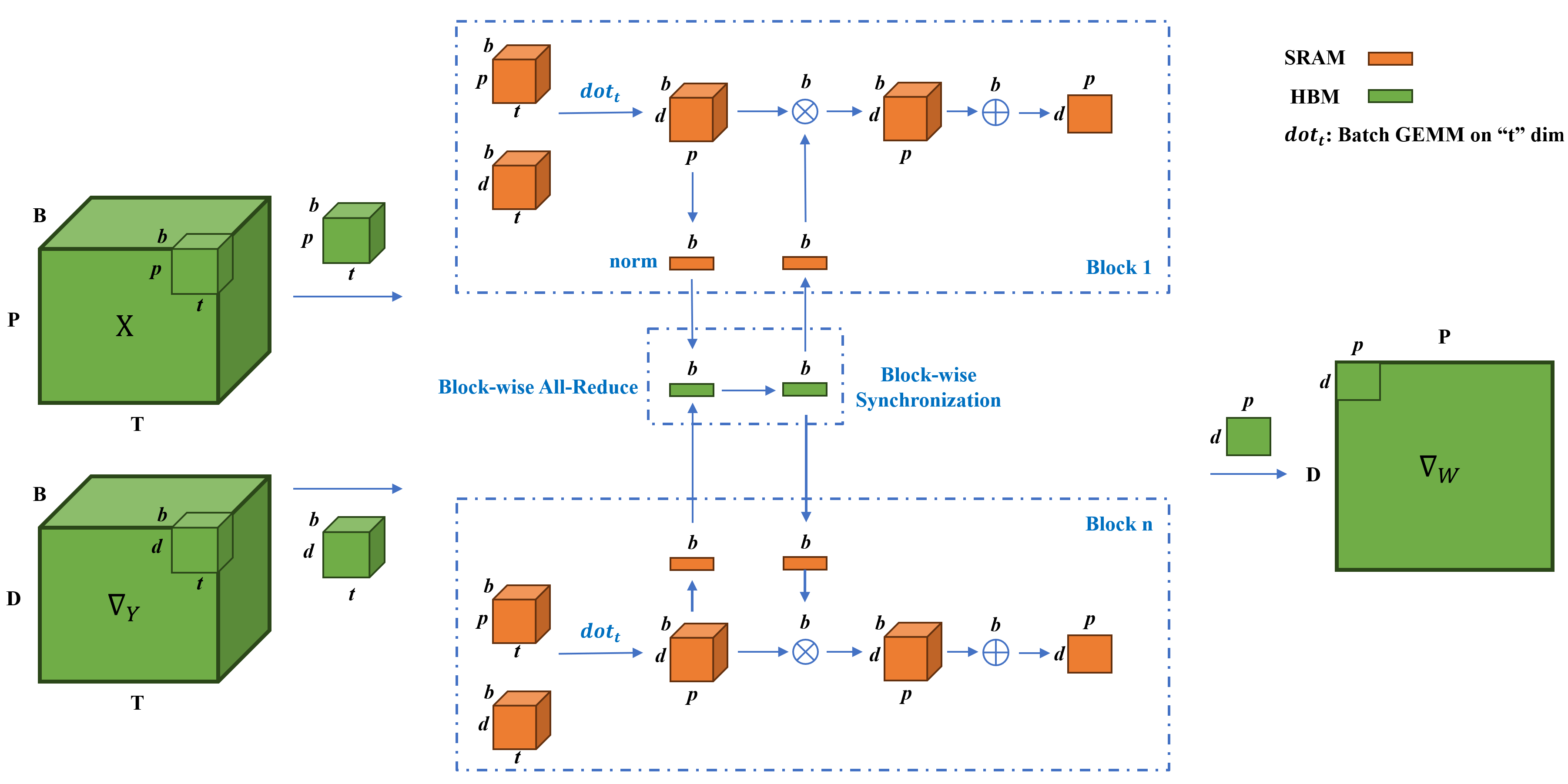}
    \caption{
    \textbf{Illustration of FlashDP.} It depicts the core algorithm design of FlashDP. Its features are integrated with on-chip per-sample gradient norm calculations. The workflow incorporates block-wise all-reduce and synchronization to facilitate efficient norm aggregation. SRAM (orange) and HBM (green) are optimally utilized to manage memory efficiently, addressing the kernel fusion challenges and reducing computational redundancy inherent in traditional DP-SGD implementations.
    }
    \label{fig:idea}
\end{figure*}

\subsection{Algorithmic Enhancements in FlashDP}

FlashDP introduces a suite of algorithmic enhancements designed to reconcile the computational demands and memory constraints associated with DP-SGD. At the heart of these enhancements is the Block-wise All-Reduce algorithm, which integrates several critical operations into a unified kernel execution, thereby optimizing on-chip memory utilization and enhancing computational throughput. 

\noindent \textbf{Efficient Kernel Fusion through Block-wise All-Reduce.} Central to FlashDP's strategy is our proposed Hierarchical Reduction Architecture (HRA), which encompasses more than just reduction operations. HRA is a structured approach that manages the computation and synchronization of data across various stages, beginning with intra-block reduction of gradient norms within individual GPU blocks. This phase employs an HRA-based reduction strategy executed in shared memory, culminating in a single norm scaler per block. Such a design significantly reduces the data footprint necessary for subsequent inter-block communications, optimizing the efficiency of the all-reduce operation across the GPU grid.

Following the compact intra-block reduction, FlashDP coordinates a global all-reduce operation across blocks, which computes a global gradient norm crucial for consistent gradient clipping across the entire mini-batch. Efficiently handled in HBM thanks to the minimized data size from earlier reductions, this step avoids the common memory bottlenecks typically associated with large-scale data operations in HBM, thus maintaining high computational throughput.

\begin{algorithm*}[h]
\renewcommand{\algorithmicrequire}{\textbf{Require:}}
\caption{Algorithm: FlashDP with Block-wise All-Reduce on GPUs}
\label{alg:flashdp}
\begin{algorithmic}[1]
\REQUIRE{Input activation tensor $X \in \mathbb{R}^{B \times T \times P}$ and output gradient tensor $\nabla_Y \in \mathbb{R}^{B \times T \times D}$ in GPU HBM}
\REQUIRE{Clipping threshold $C$, noise scale $\sigma$}
\REQUIRE Block dimensions $b$, $t$, $d$, and $p$ for batch size, sequence length, output features, and input features, respectively.
\STATE Split block for output gradient tensor $B_{\nabla_Y} \in \mathbb{R}^{b \times t \times d}$, input activation tensor $B_X \in \mathbb{R}^{b \times t \times p}$ based on GPU on-chip SRAM size $M$.
\FOR{each training backward iteration}
    \FOR{each block input index $i_p = 1, 2, \ldots, \frac{P}{p}$ in parallel}
        \FOR{each block output feature $i_d = 1, 2, \ldots, \frac{D}{d}$ in parallel}
            \FOR{each block batch size $i_b = 1, 2, \ldots, \frac{B}{b}$ in parallel}
                \STATE Load output gradient block $B_{\nabla_Y}$ and input activation block $B_X$ from HBM to SRAM.
                \STATE Compute per-sample gradients block $B_{G} = \sum_T B_{\nabla_Y}^T B_X$ on-chip SRAM.
                \STATE \textcolor{red}{Intra-block Reduce}: Compute per-sample gradients norm square block $\|B_{G}\|^2 = \sum_d \sum_p {B_{\mathbf{G}}}^2$ on-chip SRAM. 
                \STATE \textcolor{blue}{Inter-block Reduce}: Offload all per-sample gradients norm square blocks $\|B_{G}\|^2$ from SRAM to HBM, and perform block-wise all-reduce.
                \STATE \textcolor{blue}{Block-wise synchronization}: Wait until all blocks finish the all-reduce operation to get all-reduced per-sample gradients norm square blocks ${\|B_{G}\|^2}'$.
                \STATE Upload ${\|B_{G}\|^2}'$ from HBM to SRAM.
                \STATE Compute clipped per-sample gradients block $B_{G}' = B_{G} / \max\left(1, \frac{\sqrt{{\|B_{G}\|^2}'}}{C}\right)$ on-chip SRAM.
                \STATE Add noise to clipped per-sample gradients block and aggregate to compute parameter gradient block $B_{\nabla_W} = \sum_b B_{G}' + \mathcal{N}(0, \sigma^2 C^2 \mathbf{I})$ on-chip SRAM.
                \STATE Offload parameter gradient block $B_{\nabla_W}$ from SRAM to HBM.
            \ENDFOR
        \ENDFOR
    \ENDFOR
\ENDFOR
\STATE Return entire parameter gradient $\nabla_W$.
\end{algorithmic}
\end{algorithm*}

The strategic implementation of HRA not only facilitates these reductions but also orchestrates synchronized updates and data consistency across the GPU architecture. By managing data flow from the point of loading through to final computation and storage, HRA ensures that the most intensive computations are confined to the faster, on-chip memory. This methodical approach leverages the GPU's capabilities to facilitate high-performance differentially private training, minimizing memory and bandwidth overhead.

The practical implementation and operational dynamics of the FlashDP approach are thoroughly illustrated in Algorithm~\ref{alg:flashdp} and visually depicted in Figure~\ref{fig:idea}. FlashDP innovatively reduces the four distinct stages typically involved in explicit DP-SGD into a {\bf single streamlined stage}. This consolidation is achieved without adding any extra computational steps, thereby enhancing the overall efficiency of the process. 
Here is a detailed breakdown of this single streamlined stage:

\noindent {\bf Optimized Block Processing and Memory Management (Line 1-6).}
Initially, FlashDP partitions the input activation tensor \(X\) and the output gradient tensor \(\nabla_Y\) into blocks based on the SRAM capacity. This strategic partitioning is crucial for managing the limited on-chip memory more effectively and ensuring that data transfers between the HBM and SRAM are minimized.

\noindent {\bf Fused Computation of Gradients and Norms (Line 7-8).}
Within the GPU's SRAM, FlashDP simultaneously computes the per-sample gradients block and their norms square (intra-block reduce) for each block. This computation leverages the GPU's powerful batched GEMM operations, enabling it to handle large data sets efficiently. 

\noindent {\bf Block-wise All-Reduce (Line 9-11).}
After computing the gradient norms, FlashDP performs a Block-wise All-Reduce operation in parallel to aggregate these norms across all blocks (inter-block reduce). This all-reduce operation is crucial for obtaining a global view of gradient norms square, which is necessary for consistent gradient clipping across the entire batch. This step is executed efficiently within the SRAM, reducing the latency and memory bandwidth requirements typically associated with inter-GPU communications.

\noindent {\bf Per-layer Gradient Clipping and Noise Addition in SRAM (Line 12-13).}
Following the gradient and norm calculations, clipping is performed directly on the chip. Each gradient is scaled according to the computed norms and a predefined clipping threshold \(C\), ensuring compliance with DP standards. Immediately after clipping, Gaussian noise based on the noise scale \(\sigma\) and the clipping threshold is added to each gradient block. 

\noindent {\bf Efficient Parameter Aggregation (Line 14-19).}
The final step in the FlashDP algorithm involves aggregating the noisy, clipped gradients across all blocks and batches directly within SRAM. This aggregation is optimized to minimize memory accesses, ensuring that only the final gradient used for the model update is transferred back to HBM.

\subsection{Adaptive Kernel Implementation}

The implementation of the FlashDP algorithm leverages the robust and versatile capabilities of the PyTorch framework \citep{paszke2019pytorch}, which is renowned for its intuitive handling of automatic differentiation and dynamic computational graphs. One of the critical features of our implementation involves customizing PyTorch's autograd functionality to accommodate the specific needs of differential privacy during the training of deep neural networks. To this end, operators that necessitate trainable parameters are intricately defined by wrapping them within PyTorch’s autograd function.

However, implementing the Block-wise All-Reduce algorithm has presented unique challenges, primarily due to the limitations of CUDA's programming model in facilitating block-wise synchronization. Block-wise synchronization is essential in our algorithm; without it, clip operations might be executed prematurely, while the inter-block reduce operation is still incomplete, leading to numerical inaccuracies in the computation of per-sample gradients' norm squares. There are two primary methods to implement synchronization: 1. cooperative groups (CG) \footnote{https://developer.nvidia.com/blog/cooperative-groups/} and 2. adaptive kernel. We opted for the second method because the grid synchronization required by CG necessitates launching all blocks simultaneously, which is impractical for DP applications.

To address this limitation, FlashDP's implementation employs an adaptive approach. Instead of relying on a monolithic kernel to perform the entire Block-wise All-Reduce operation, the process is split across different kernels, which are executed iteratively over the batch dimension. This iterative approach allows for synchronization points between the execution of kernels, using the inherent block synchronization that occurs at kernel launch and completion.

The execution flow in FlashDP is as follows: (1) \textbf{Intra-block Reduction}: Each block computes the norms of its gradients and performs an HRA-based reduction within the shared memory. This step employs a shuffle-reduce mechanism, optimizing intra-block operations by minimizing memory footprint and synchronization overhead. This results in a single norm value per block. (2) \textbf{Inter-block Reduction}: Each block transfers the outcome of its intra-block reduction to the HBM. This transfer is facilitated through atomic operations for several reasons. Firstly, the result of the intra-block reduction comprises only a single element, and each block elects only one thread to perform the atomic operation on this element. This approach minimizes potential bottlenecks, as the differing execution speeds across blocks prevent serious serialization issues. Secondly, atomic operations benefit from acceleration by the hardware instruction set, ensuring that these operations are executed swiftly and efficiently. (3) \textbf{Inter-kernel Synchronization}: After the completion of the inter-block reduction, FlashDP leverages the termination of the kernel as a natural synchronization point. At this juncture, all blocks have finished their individual reductions. (4) \textbf{Iterative Kernel Launch}: For each batch element, a new kernel is launched serially, maintaining synchronization across kernels. This approach involves broadcasting operations where source operands are dimensionally disparate, ensuring uniform data handling across computational units. 

This implementation strategy, while divergent from the ideal single-kernel solution, allows FlashDP to function effectively within the current constraints of CUDA. It underscores FlashDP's adaptability and represents a practical solution to the block synchronization challenge, ensuring accurate gradient norm calculations essential for maintaining the model's differential privacy. 

\section{Experiments}

Our experimental suite is methodically designed to assess the robustness and efficiency of FlashDP across a range of training paradigms and hardware configurations. We explore FlashDP's performance in terms of memory efficiency and throughput under varying batch sizes, its adaptability to Automatic Mixed Precision (AMP) training (Appendix Section \ref{sec:exp-amp}), its scalability when employing Distributed Data Parallel (DDP) and Pipeline Parallel (PP) techniques (Appendix Section \ref{sec:exp-dist}), and utility evaluation (Appendix Section \ref{sec:utility}).

\begin{table*}[!h]
\centering
\caption{\textbf{Differential Batch-size Analysis.} The table displays a multi-panel comparison of memory usage and throughput for four differential privacy methods--NonDP, Opacus, GhostClip, BK, and FlashDP--across different batch sizes B (1, 2, 4, and 8) when applied to GPT-2 models of varying sizes (small, medium, and large). Instances of `-' in the table indicate scenarios where the corresponding method failed to execute due to memory constraints.}
\resizebox{\textwidth}{!}{
\begin{tabular}{ccccccc|ccccc}
\hline
\multirow{2}{*}{Model} & \multirow{2}{*}{B} & \multicolumn{5}{c}{Memory Usage (MB x1e4)} & \multicolumn{5}{c}{Throughput (tokens/sec x1e4)} \\ \cline{3-12} 
 &  & NonDP  & Opacus  & GhostClip  & BK  & FlashDP  & NonDP  & Opacus  & GhostClip  & BK  & FlashDP  \\ \hline
GPT2-small &  & 0.50 & 0.75(x1.50) & {\bf 0.46(x0.92)} & 0.53(x1.06) & 0.50(x1.00) & 2.84 & 0.91(x0.32) & 0.57(x0.20) & 1.56(x0.54) & {\bf 1.83(x0.64)} \\
GPT2-medium & 1 & 1.26 & 1.53(x1.21) & {\bf 1.12(x0.89)} & 1.68(x1.33) & 1.26(x1.00) & 1.10 & 0.42(x0.38) & 0.39(x0.35) & 0.75(x0.68) & {\bf 0.86(x0.78)} \\
GPT2-large &  & 2.48 & 3.99(x1.61) & {\bf 2.17(x0.88)} & 2.73(x1.18) & 2.48(x1.00) & 0.58 & 0.25(x0.43) & 0.27(x0.46) & 0.40(x0.69) & {\bf 0.51(x0.89)} \\ \hline
GPT2-small &  & 0.87 & 1.30(x1.49) & {\bf 0.79(x0.91)} & 1.01(x1.16) & 0.87(x1.00) & 3.22 & 1.68(x0.52) & 0.92(x0.29) & 1.91(x0.59) & {\bf 2.32(x0.72)} \\
GPT2-medium & 2 & 2.07 & 2.89(x1.39) & {\bf 1.87(x0.90)} & 2.44(x1.18) & 2.07(x1.00) & 1.28 & 0.74(x0.58) & 0.59(x0.46) & 0.81(x0.63) & {\bf 1.02(x0.80)} \\
GPT2-large &  & 3.91 & 4.79(x1.23) & {\bf 3.53(x0.90)} & 4.81(x1.23) & 3.91(x1.00) & 0.68 & 0.38(x0.56) & 0.38(x0.56) & 0.45(x0.66) & {\bf 0.59(x0.87)} \\ \hline
GPT2-small &  & 1.53 & 2.07(x1.35) & {\bf 1.44(x0.94)} & 1.68(x1.09) & 1.53(x1.00) & 3.60 & 2.42(x0.67) & 1.42(x0.39) & 2.24(x0.62) & {\bf 2.59(x0.72)} \\
GPT2-medium & 4 & 3.58 & 4.26(x1.19) & {\bf 3.33(x0.93)} & 4.00(x1.12) & 3.58(x1.00) & 1.42 & 0.90(x0.63) & 0.81(x0.57) & 0.95(x0.67) & {\bf 1.13(x0.80)} \\
GPT2-large &  & 6.60 & - & {\bf 6.15(x0.93)} & 6.60(x1.00) & 6.60(x1.00) & 0.76 & - & 0.50(x0.66) & 0.53(x0.70) & {\bf 0.64(x0.84)} \\ \hline
GPT2-small &  & 2.86 & 3.44(x1.20) & {\bf 2.72(x0.95)} & 2.86(x1.00) & 2.86(x1.00) & 3.80 & 2.64(x0.69) & 1.92(x0.51) & 2.40(x0.63) & {\bf 2.72(x0.72)} \\
GPT2-medium & 8 & 6.60 & - & {\bf 6.24(x0.95)} & 6.60(x1.00) & 6.60(x1.00) & 1.52 & - & 0.99(x0.65) & 1.03(x0.68) & {\bf 1.19(x0.78)} \\
GPT2-large &  & - & - & - & - & - & - & - & - &- &- \\ \hline
\end{tabular}
}
\label{tab:result-batchsize}
\end{table*}

\subsection{Experimental Setup}

Our experiments utilize the Wikitext dataset \citep{merity2016wikitext} and are conducted on NVIDIA A100 (80GB) GPUs using the PyTorch framework \citep{paszke2019pytorch}. We assess the performance of FlashDP across various configurations by comparing it with established explicit methods Opacus \citep{yousefpour2021opacus}, and implicit method GhostClip \citep{li2021large} and BK \citep{bu2023accuracy}, all in the per-layer clipping mode, under different training paradigms.\footnote{As \citep{bu2023accuracy,he2022exploring} demonstrated that for LLMs, compared to global clipping, per-layer clipping is more memory-efficient and time-efficient while achieving comparable performance in terms of both privacy preservation and accuracy. Here, we only consider per-layer clipping baselines.} The tested models include GPT-2 \citep{radford2019language} with a sequence length of 1024 and Llama \citep{touvron2023llama} models, both with a sequence length of 2048. 
More experimental settings and explanations can be found in Appendix~\ref{sec:exp-settings}.

\subsection{Results of Batch Size \& Micro Batch Size}

Efficient batch processing is crucial in LLM training due to its high computational and memory demands. By examining both batch and micro-batch sizes, we assess FlashDP’s ability to manage memory more effectively and maintain high throughput. This also tests the practicality of gradient accumulation (GA), which allows larger effective batch sizes by splitting them into smaller, manageable micro-batches. The experiment results of different micro batch sizes can be seen in Appendix \ref{sec:exp-micro-batch-size}.

In Table~\ref{tab:result-batchsize}, FlashDP was benchmarked against traditional DP-SGD methods like Opacus, GhostClip, and BK, as well as a non-DP (NonDP) configuration, demonstrating superior memory efficiency and throughput. FlashDP utilized approximately 38\% less memory than Opacus and nearly matched the NonDP configuration while processing the GPT-2 large model at a batch size of 1. It achieved a throughput nearly double that of Opacus and only slightly lower than NonDP, showcasing its effective balance between privacy preservation and computational efficiency. Opacus exhibited the highest memory usage, which escalated with batch size, leading to failure at a batch size of 8. GhostClip, while more memory-efficient than Opacus, suffered from reduced throughput at higher batch sizes due to gradient re-computation. BK's performance was intermediate, lacking distinct advantages. Overall, FlashDP not only maintained lower memory usage and higher throughput than the DP methods across all batch sizes but also approached the efficiency of NonDP configurations.

\subsection{Results of Distributed Training}
\label{sec:exp-dist}

Distributed Data Parallel (DDP) \citep{li2020pytorch} and Pipeline Parallel (PP) \citep{kim2020torchgpipe} are two advanced techniques crucial for scaling the training of LLMs efficiently across multiple GPUs or nodes. 

\begin{figure*}[ht]
    \centering
    \subfigure[Memory Usage]{
        \begin{minipage}{.37\textwidth}  
            \includegraphics[width=\textwidth]{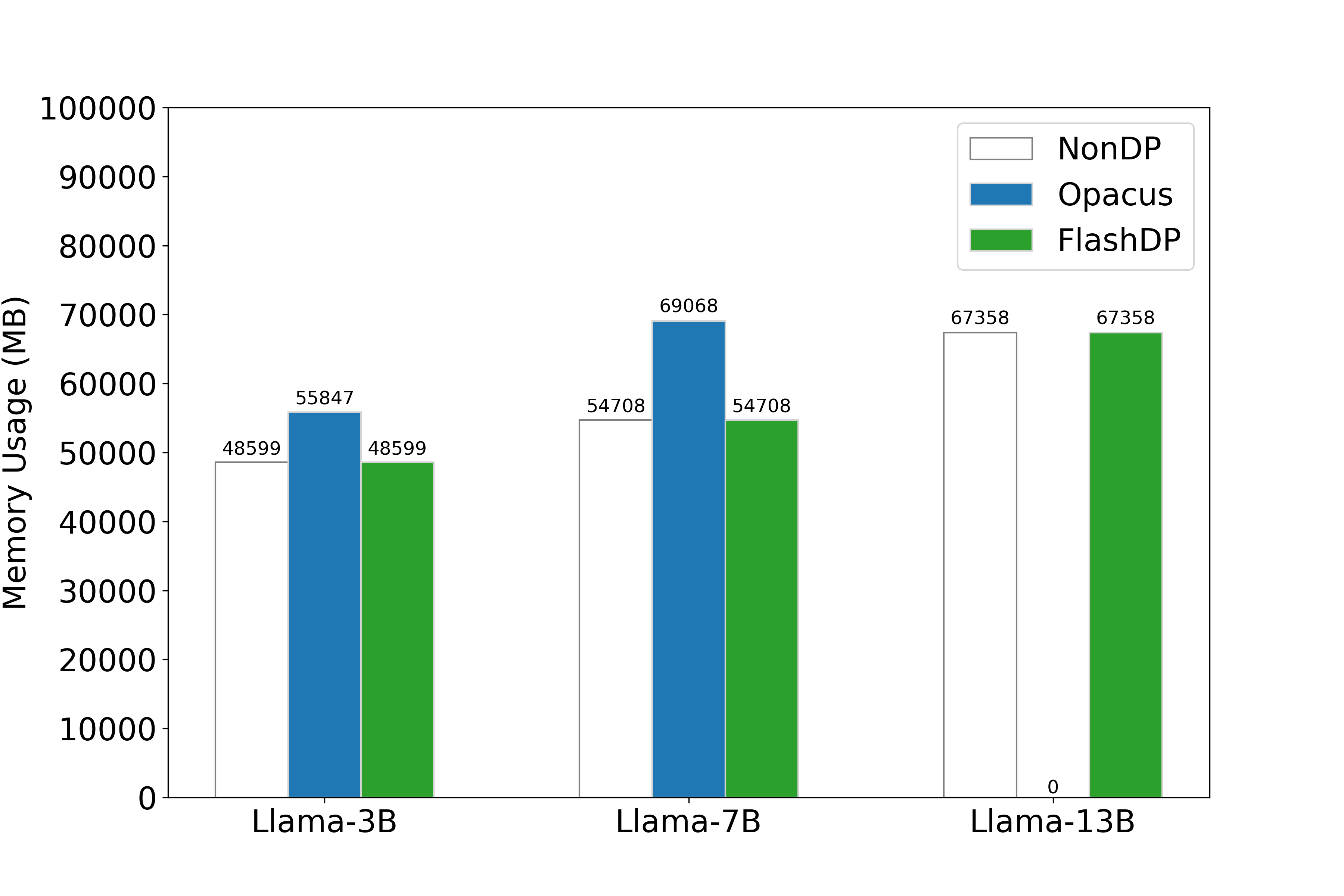}
                \label{fig:memory}
        \end{minipage}
    }
    \subfigure[Throughput]{
        \begin{minipage}{.37\textwidth}
            \includegraphics[width=\textwidth]{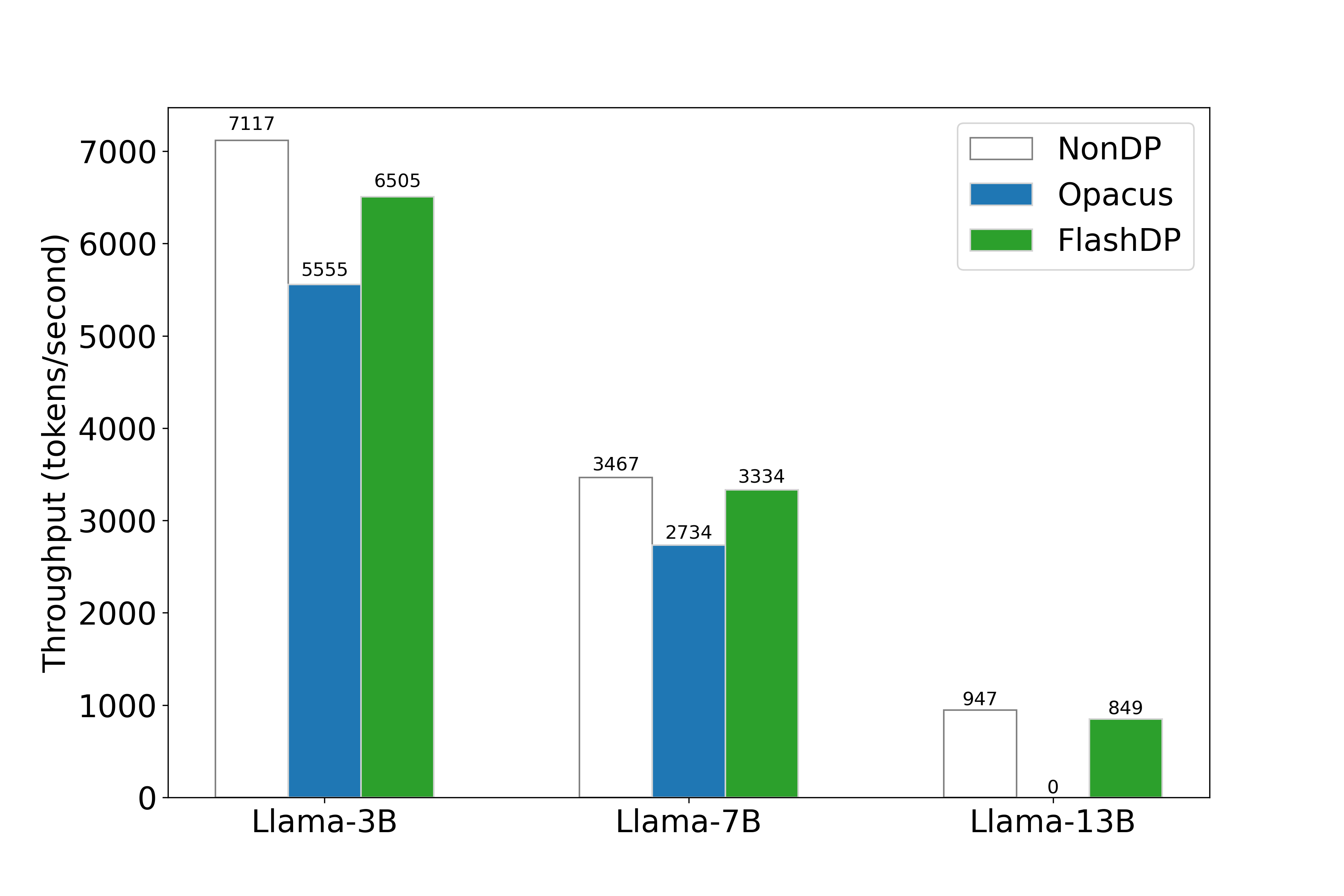}
                \label{fig:throughput}
        \end{minipage}
    }
    \caption{
    \textbf{Memory and Throughput for Llama Models Using Pipeline Parallel Training.} (a) Memory usage for Llama-3B, Llama-7B, and Llama-13B models. (b) Throughput in tokens per second across these model sizes. A value of 0 indicates out of memory.
    }
    \label{fig:result-pp}
\end{figure*}

\noindent {\bf Distributed Data Parallel (DDP).}
Figure \ref{fig:result-ddp} in Appendix illustrates the performance of different methods in a DDP setting across GPT-2 models of varying sizes. FlashDP showcases superior memory usage efficiency and higher throughput across all model sizes when compared to Opacus and BK. Notably, even as the model size increases, FlashDP maintains a competitive edge close to the NonDP benchmarks, highlighting its effective parameter distribution and gradient computation across multiple GPUs. This is crucial in scenarios where training speed and model scalability are priorities.

\noindent {\bf Pipeline Parallel (PP).}
In the PP scenario depicted in Figure \ref{fig:result-pp}, FlashDP was tested with Llama models varying from 3 billion to 13 billion parameters. The results indicate that FlashDP not only scales efficiently with increasing model size but also demonstrates significant throughput improvements compared to Opacus and BK. Particularly, FlashDP’s ability to handle the largest model (Llama-13B) with minimal throughput degradation illustrates its robustness in managing extensive computational loads, characteristic of PP environments.

\section{Conclusion}
In this paper, we introduce FlashDP, a cache-friendly approach to per-layer DP-SGD that improves memory efficiency and computational throughput for large language model (LLM) training. By optimizing GPU I/O through a unified Block-wise All-Reduce algorithm and a Hierarchical Reduction Architecture (HRA), FlashDP significantly reduces memory transactions and eliminates redundant computations. We also adopt an adaptive kernel design to overcome CUDA's synchronization limitations. Experiments show that FlashDP achieves memory usage close to non-private baselines and maintains 90\% throughput during Llama-13B training on four A100s, without compromising privacy or accuracy.
FlashDP may enable the deployment of privacy-preserving LLMs in sensitive domains such as healthcare, education, and finance, where data protection is critical. At the same time, it highlights the need for responsible release practices to mitigate potential misuse under the guise of privacy.

\bibliography{main}
\bibliographystyle{cite}

\appendix

\section{Preliminaries}\label{sec:pre}
\subsection{Differential Privacy}
\label{sec:background-dp}

\begin{definition}
    \label{def:dp}
    (Differential Privacy \citep{dwork2006calibrating})
    Given a data universe $\mathcal{X}$, two datasets $X, X'\subseteq \mathcal{X}$ are adjacent if they differ by one data example. A randomized algorithm $\mathcal{M}$ is $(\varepsilon,\delta)$-differentially private if for all adjacent datasets $X,\, X'$ and for all events $S$ in the output space of $\mathcal{M}$, we have $\operatorname{Pr}(\mathcal{M}(X)\in S)\leq e^{\varepsilon} \operatorname{Pr}(\mathcal{M}(X')\in S)+\delta$.
\end{definition}

\noindent\textbf{Differentially Private Stochastic Gradient Descent (DP-SGD) \citep{ abadi2016deep}.} DP-SGD is an adaptation of this principle for machine learning models, where privacy is preserved during the training process by modifying the gradient computation.

In the context of a model parameterized by weights \(\theta\) for loss $\mathcal{L}$, the standard SGD update is modified in DP-SGD to include a mechanism for privacy preservation. Specifically, the gradient \(\nabla \mathcal{L}(\theta, x_i)\) for each training example \(x_i\) is first computed, and then processed as follows to incorporate privacy:

\begin{enumerate}
    \item \textbf{Clipping:} Each gradient is clipped to a maximum norm \(C\), defined as:
    $g'_i = {g_i}{\min(1, \frac{C}{\|g_i\|_2})},$ 
    where \(g_i = \nabla \mathcal{L}(\theta, x_i)\). 

    \item \textbf{Noise Addition:} Gaussian noise is added to the aggregated clipped gradients to ensure differential privacy:
    \[
    \tilde{g} = \frac{1}{B} \sum_{i=1}^B g'_i + \mathcal{N}(0, \sigma^2 C^2 I)
    \]
    where \(B\) is the batch size, and \(\sigma\) is the noise scale, determined by the privacy budget, subsampling rate, and iteration number.
\end{enumerate}

The model parameters are then updated using the noisy, aggregated gradient:
\(
\theta \leftarrow \theta - \eta \tilde{g}, 
\)
where \(\eta\) is the learning rate. This approach to privacy-preserving training addresses the fundamental trade-off between accuracy and privacy by controlling the granularity of the updates through the parameters \(C\) and \(\sigma\).

In this work, we actually use Per-layer clipping and Differentially Private Adam (DP-Adam) instead of standard DP-SGD. 
The key distinction of the per-layer DP-SGD compared to standard DP-SGD lies in its approach to clipping gradients layer by layer and incorporating noise accordingly. (While there are various adaptations of per-layer DP-SGD, we focus on the simplest format that directly extends from the standard DP-SGD.) \citep{bu2023accuracy} have demonstrated that per-layer clipping not only matches the accuracy of global clipping but also significantly enhances memory and throughput efficiency.
While DP-Adam incorporates the same mechanisms for gradient clipping and noise addition as described for DP-SGD, it also leverages the adaptive learning rates characteristic of Adam. The detailed algorithms can be found in Algorithm \ref{alg:dp-grad}-\ref{alg:per-layer-dp-grad}-\ref{alg:dpadam}. 

\begin{algorithm}[h]
\renewcommand{\algorithmicrequire}{\textbf{Require:}}
\caption{Common Gradient Processing in DP-SGD and DP-Adam}
\label{alg:dp-grad}
\begin{algorithmic}[1]
\REQUIRE $\mathcal{L}(\theta, x_i)$: Loss function for parameter \(\theta\) and input \(x_i\)
\REQUIRE $C$: Clipping threshold
\REQUIRE $\sigma$: Noise scale
\REQUIRE $B$: Batch size
\FOR{$i = 1$ to $B$}
    \STATE Compute gradient: \( g_i = \nabla \mathcal{L}(\theta, x_i) \)
    \STATE Clip gradient: \( g'_i = g_i \min(1, \frac{C}{\|g_i\|_2}) \)
\ENDFOR
\STATE Aggregate clipped gradients and add Gaussian noise: \( \tilde{g} = \frac{1}{B} \sum_{i=1}^B g'_i + \mathcal{N}(0, \sigma^2 C^2 I) \)
\end{algorithmic}
\end{algorithm}

\begin{algorithm}[h]
\renewcommand{\algorithmicrequire}{\textbf{Require:}} 
\caption{Per-Layer Gradient Processing in DP-SGD} 
\label{alg:per-layer-dp-grad} 
\begin{algorithmic}[1] 
\REQUIRE $\mathcal{L}(\theta^{(l)}, x_i)$: Loss function for layer parameters \(\theta^{(l)}\) and input \(x_i\) 
\REQUIRE $C^{(l)}$: Clipping threshold for layer \(l\) 
\REQUIRE $\sigma^{(l)}$: Noise scale for layer \(l\) 
\REQUIRE $L$: Total number of layers 
\FOR{each layer $l = 1$ to $L$} 
    \FOR{$i = 1$ to $B$} 
        \STATE Compute gradient for layer \(l\): \( g^{(l)}_i = \nabla \mathcal{L}(\theta^{(l)}, x_i) \) 
        \STATE Clip gradient for layer \(l\): \( g'^{(l)}_i = g^{(l)}_i \min\left(1, \frac{C^{(l)}}{\|g^{(l)}_i\|_2}\right) \) 
    \ENDFOR 
    \STATE Aggregate clipped gradients for layer \(l\) and add Gaussian noise: \( \tilde{g}^{(l)} = \frac{1}{B} \sum_{i=1}^B g'^{(l)}_i + \mathcal{N}(0, (\sigma^{(l)} C^{(l)})^2 I) \) 
\ENDFOR 
\end{algorithmic} 
\end{algorithm}

\begin{algorithm}[h]
\renewcommand{\algorithmicrequire}{\textbf{Require:}}
\caption{DP-SGD Specific Steps}
\begin{algorithmic}[1]
\REQUIRE $\theta$: Model parameters
\REQUIRE $\eta$: Learning rate
\FOR{each training step}
    \STATE Perform common gradient processing as in Algorithm~\ref{alg:dp-grad}
    \STATE Update model parameters: \( \theta \leftarrow \theta - \eta \tilde{g} \)
\ENDFOR
\end{algorithmic}
\end{algorithm}

\begin{algorithm}[h]
\renewcommand{\algorithmicrequire}{\textbf{Require:}}
\caption{DP-Adam Specific Steps}
\label{alg:dpadam}
\begin{algorithmic}[1]
\REQUIRE $m, v$: Estimates of the first and second moments (initially 0)
\FOR{each training step}
    \STATE Perform common gradient processing as in Algorithm~\ref{alg:dp-grad}
    \STATE Update moment estimates: \( m \leftarrow \beta_1 m + (1 - \beta_1) \tilde{g} \)
    \STATE \( v \leftarrow \beta_2 v + (1 - \beta_2) \tilde{g}^2 \)
    \STATE Compute adaptive learning rate: \( \hat{\eta} = \eta / (\sqrt{v} + \epsilon) \)
    \STATE Update parameters: \( \theta \leftarrow \theta - \hat{\eta} m \)
\ENDFOR
\end{algorithmic}
\end{algorithm}

\subsection{Transformers}
\label{sec:background-transformers}

The transformer architecture, proposed by Vaswani et al.~\citep{vaswani2017attention}, is predicated on self-attention mechanisms that process input tokens in parallel, significantly improving the performance and training efficiency of sequence-to-sequence tasks. This architecture has become the backbone of LLMs. 

In a transformer model, the input tensor $\mathbf{X}$  of size $B \times T \times P$ (since we are considering LLM, so we only focus on text data as the input), where $B$ is the batch size, $T$ is the sequence length (number of tokens), and $P$ is the embedding size of a token, undergoes a series of transformations through multi-head self-attention and feedforward neural network blocks. For each token in the sequence, the transformer computes a weighted sum of all tokens in the input, where the weights are determined through the self-attention mechanism.

\noindent \textbf{Multi-Head Attention (MHA).} The attention mechanism is primarily built upon linear transformations where the query $\mathbf{Q}$, key $\mathbf{K}$, and value $\mathbf{V}$ matrices are obtained as follows:
\begin{equation}
\mathbf{Q} = \mathbf{X}\mathbf{W}_Q, \quad \mathbf{K} = \mathbf{X}\mathbf{W}_K, \quad \mathbf{V} = \mathbf{X}\mathbf{W}_V
\end{equation}
where $\mathbf{W}_Q$, $\mathbf{W}_K$, and $\mathbf{W}_V$ are the weight matrices that are subject to training.

\noindent \textbf{Feedforward Network (FFN).} The FFN in the transformer  consists of two linear transformations with a ReLU activation in between:
\begin{equation}
\text{FFN}(\mathbf{x}) = \text{ReLU}(\mathbf{x}\mathbf{W}_1)\mathbf{W}_2
\end{equation}
Here, $\mathbf{W}_1$ and $\mathbf{W}_2$ are the weight matrices, all of which are trainable parameters of the linear layers within the FFN.

\noindent \textbf{Layer Normalization (LN).} LN is applied post-attention and FFN in each layer of the transformer. It normalizes the output of each neuron to have a mean of zero and a variance of one, which are then scaled and shifted by the trainable parameter vectors $\boldsymbol{\gamma}$ and $\boldsymbol{\beta}$, respectively:
\begin{equation}
\text{LayerNorm}(\mathbf{x}) = \boldsymbol{\gamma} \odot \left(\frac{\mathbf{x} - \mu}{\sqrt{\sigma^2 + \epsilon}}\right) + \boldsymbol{\beta}
\end{equation}
where $\mu$ and $\sigma^2$ are the mean and variance calculated over the last dimension of the input tensor $\mathbf{x}$, $\epsilon$ is a small constant added for numerical stability, and $\odot$ denotes element-wise multiplication. The layer normalization parameters $\boldsymbol{\gamma}$ (scale) and $\boldsymbol{\beta}$ (shift) are learned to optimally scale and shift the normalized data.

The key trainable parameters in the transformer model are:
\begin{enumerate}
    \item Weights of the WHA mechanism, including query $\mathbf{W}_Q$, key $\mathbf{W}_K$, and value $\mathbf{W}_V$ matrices, each of size $P \times P$.
    \item Position-wise FFN weights $\mathbf{W}_1$ of size $P \times H$ and $\mathbf{W}_2$ of size $H \times P$, where $H$ is the hidden layer size.
    \item LN parameters $\boldsymbol{\gamma}$ and $\boldsymbol{\beta}$, which are vectors of size $P$.
\end{enumerate}

It is important to highlight that the bulk of the trainable parameters in the transformer model stems from MHA and FFN modules, both of which consist of linear transformations. These linear parameters are responsible for the vast majority of transformations within the transformer and significantly contribute to its parameter count. In contrast, the trainable parameters in LN represent a relatively smaller portion of the model’s total parameters. Therefore, we focus on the linear parameters gradient computation. 

\noindent \textbf{DP-SGD for Training Transformers.} 
The process of adapting DP-SGD to transformers is formalized as follows:
For each batch of input data $X$ and corresponding loss function $\mathcal{L}$, compute the per-sample gradients $\mathbf{G}_\theta$ for all trainable parameters $\theta = \{\mathbf{W}_Q, \mathbf{W}_K, \mathbf{W}_V, \mathbf{W}_1, \mathbf{W}_2, \boldsymbol{\gamma}, \boldsymbol{\beta}\}$:
\begin{equation}
\mathbf{G}_\theta = \nabla_{\theta} \mathcal{L}(\theta, X) \in \mathbb{R}^{B \times |\theta|}. 
\end{equation}
where  $\nabla_{\theta} \mathcal{L}(\theta, X)$ denotes the computation of gradients of the loss with respect to the parameters $\theta$ for the batch $X$.

\subsection{GPU Architecture and CUDA Programming}
\label{sec:background-cuda}

High performance in deep learning, particularly in operations like General Matrix to Matrix Multiplication (GEMM), is largely attributable to the parallel processing power of modern Graphics Processing Units (GPUs). The architectural design of GPUs, with their numerous cores and hierarchical memory systems, is optimized for the parallel execution of operations, making them ideal for the matrix-intensive computations required in neural network training.

\noindent \textbf{GPU Architecture.} At the heart of GPU's computational efficiency are its Streaming Multiprocessors (SMs), which are essentially multiprocessor units that execute a large number of threads concurrently. Each SM is a powerhouse of performance, containing a set of processing cores and a block of on-chip memory, primarily Shared Random Access Memory (SRAM), which includes registers and shared memory. Shared memory, an ultra-fast SRAM, allows threads within the same block to exchange data without involving the slower global memory (HBM), thus acting as a crucial facilitator for matrix blocking.

\noindent \textbf{CUDA and GEMM.} The quintessential challenge in optimizing GEMM lies in the meticulous orchestration of data movement and computation, an endeavor where matrix blocking emerges as a pivotal strategy. Leveraging the robust architecture of GPUs and the sophisticated abstractions provided by CUDA (Compute Unified Device Architecture), matrix blocking transforms the theoretical prowess of parallel computation into a practical performance paradigm.

\noindent \textbf{Principles of Matrix Blocking.} Matrix blocking, also known as matrix tiling, is a technique ingeniously conceived to enhance data locality and parallelism. It systematically partitions extensive matrix operands into smaller, manageable sub-matrices or 'blocks' that can be independently dispatched to the GPU’s SMs. The judicious use of shared memory within SMs for these blocks reduces the frequency and volume of global memory accesses, a common bottleneck due to its higher latency. Blocking is pivotal in minimizing the communication overhead between the slow global memory and the fast but limited on-chip shared memory. This stratagem leverages the temporal and spatial locality by reusing data within the fast-access memory hierarchies, significantly reducing the volume of data shuttled to and from the global memory, thereby enhancing the computational throughput.

\noindent  \textbf{Mathematical Formalization of Blocking GEMM.} Consider the GEMM operation defined as $\mathbf{C} = \mathbf{A} \times \mathbf{B}$, where $\mathbf{A} \in \mathbb{R}^{m \times n}$, $\mathbf{B} \in \mathbb{R}^{n \times p}$, and the resultant matrix $\mathbf{C} \in \mathbb{R}^{m \times p}$. Blocking decomposes this operation into smaller, tractable computations over blocks such that:
\begin{equation}
\mathbf{C}_{ij} = \sum_{k=1}^{N} \mathbf{A}_{ik} \times \mathbf{B}_{kj},
\end{equation}
where $N$ is the number of blocks, and each $\mathbf{C}_{ij}$, $\mathbf{A}_{ik}$, and $\mathbf{B}_{kj}$ represents a sub-matrix or block within $\mathbf{C}$, $\mathbf{A}$, and $\mathbf{B}$, respectively. The indices $i$, $j$, and $k$ denote the specific block within the partitioned matrices.

The dimensions of each block are chosen based on the GPU's shared memory constraints and the size of the SMs’ thread blocks, enabling optimal utilization of resources. These dimensions are represented as $B_{m} \times B_{n}$ for $\mathbf{A}_{ik}$ and $B_{n} \times B_{p}$ for $\mathbf{B}_{kj}$, leading to a block $\mathbf{B}_C$ in size of $B_{m} \times B_{p}$ for $\mathbf{C}_{ij}$. Hence, the computational paradigm shifts to:
\begin{equation}
\mathbf{B}_{C_{ij}} = \sum_{k=1}^{B_{n}} (\mathbf{B}_{A_{ik}} \times \mathbf{B}_{B_{kj}}),
\end{equation}
where each multiplication within the summation is an independent block-level GEMM that can be executed in parallel.

\iffalse 
\subsection{Gradient Computation in DP-SGD and DP-Adam}
\label{apx:grad}

Both Differentially Private Stochastic Gradient Descent (DP-SGD) and Differentially Private Adam (DP-Adam) incorporate core mechanisms to ensure differential privacy by modifying the gradient computation process. The shared processing steps include gradient clipping and noise addition, which are crucial for maintaining privacy during training.

\subsubsection{Shared Gradient Processing Steps}
The gradient computation for both DP-SGD and DP-Adam can be summarized as follows:

\subsubsection{Specifics of DP-SGD}
DP-SGD utilizes the gradient processing framework established in Algorithm~\ref{alg:dp-grad}, with specific adaptations to the standard stochastic gradient descent process to accommodate privacy concerns.

\subsubsection{Specifics of DP-Adam}
While DP-Adam follows the same gradient clipping and noise addition protocol as DP-SGD, it incorporates additional features from the Adam optimizer. Specifically, DP-Adam uses adaptive learning rates based on estimates of first and second moments of the processed gradients.

This presentation explicitly indicates that DP-SGD and DP-Adam share the initial steps of gradient computation (clipping and noise addition).

\fi 

\section{Details of Training Workflow}\label{sec:work_all}

\subsection{Non-private Training Workflow}
\label{sec:non-private-workflow}

In the standard training regime without privacy constraints, the linear forward operation takes an activation tensor $X \in \mathbb{R}^{B \times T \times P}$ and a weight matrix $W \in \mathbb{R}^{D \times P}$, producing an output $Y \in \mathbb{R}^{B \times T \times D}$ according to the matrix multiplication $Y = XW^{\mathsf{T}}$, where B, T, P, and D indicate the batch size, sequence length (token length), feature dimension of input activation tensor $X$, and feature dimension of output activation tensor $Y$, respectively. 

During the backward pass, the gradient of the output with respect to the loss, denoted by $\nabla_Y \in \mathbb{R}^{B \times T \times D}$, is computed to be of the same dimensions as the output tensor $Y$. Subsequently, the gradient with respect to the weight matrix $W$, denoted by $\nabla_W \in \mathbb{R}^{D \times P}$, is obtained by summing the product of the transpose of the gradient tensor of each batch item and the corresponding input tensor, expressed as $\nabla_W = \sum_{B} \sum_{T} (\nabla_Y)^{\mathsf{T}} X$, where $\sum_{B}$ represents the summation along the dimension $B$ (similar for other notations). 

Figure~\ref{fig:compare} (a) illustrates the computational workflow for the forward and backward pass of a linear operation within this conventional training framework. As shown in the figure, the activation tensor $X$ and the weights $W$ reside in HBM, which allows for rapid parallel access and is typically used for storing larger datasets and model parameters during GPU computations. The intermediate dot products and summations are handled using SRAM, shown in orange, which is faster than HBM and suitable for storing temporary, small blocks of data during computation. This setup minimizes memory access time and maximizes throughput.

\paragraph{Clarification on Gradient Formulations and Reviewer Feedback.}
To clarify the gradient computation in our NonDP baseline and address a reviewer's concern, we now present two mathematically equivalent formulations:

\textbf{Format 1 (Used in Our Implementation).} The default in frameworks like PyTorch is to use batched matrix operations. Let $\nabla_Y \in \mathbb{R}^{B \times T \times D}$ be the gradient of the output and $X \in \mathbb{R}^{B \times T \times P}$ be the input activation. The weight gradient $\nabla_W \in \mathbb{R}^{D \times P}$ is computed as:
\begin{equation}
\nabla_W = (\nabla_Y)^{\mathsf{T}} \cdot X
\label{eq:batched-grad}
\end{equation}
This corresponds to the batched GEMM routine invoked during \texttt{loss.backward()} and does not require computing or storing per-sample gradients.

\textbf{Format 2 (Shown in Figure~\ref{fig:compare}a for Comparison Only).}  
For structural alignment with the DP workflows, we also illustrate an equivalent formulation that computes per-sample gradients and then aggregates them:
\begin{equation}
G^{(b)} = \sum_{t=1}^{T} (\nabla_Y^{(b,t)})^{\mathsf{T}} X^{(b,t)}, \quad
\nabla_W = \sum_{b=1}^{B} G^{(b)}
\label{eq:per-sample-grad}
\end{equation}
This version is shown in Figure~\ref{fig:compare} (a) only to highlight architectural differences across methods. We reiterate that this is not used in our actual NonDP implementation.

\textbf{Summary.} We emphasize that our implementation of the NonDP baseline strictly uses Format~\ref{eq:batched-grad} (batched GEMM) and does not compute or store per-sample gradients. The inclusion of per-sample nodes in Figure~\ref{fig:compare}(a) is purely illustrative and will be clarified in the revised caption and main text.

\subsection{Explicit DP-SGD Workflow}
\label{sec:explicit-workflow}

Figure \ref{fig:compare} (b) terms the explicit method (e.g., Opacus, FastClip), demonstrates the traditional DP approach where per-sample gradients are stored explicitly, resulting in increased memory usage due to the retention of individual gradient information for noise addition and clipping.
The explicit DP-SGD workflow is normally organized into four distinct stages to ensure adherence to privacy constraints:

\textbf{Stage 1: Per-sample Gradient Computation.} At this initial stage, the activation tensor $X \in \mathbb{R}^{B \times T \times P}$ and the output gradient tensor $\nabla_Y \in \mathbb{R}^{B \times T \times D}$ are loaded in blocks from the HBM to the on-chip SRAM. The per-sample gradients tensor $\mathbf{G} \in \mathbb{R}^{B \times D \times P}$ is computed by performing the operation $\mathbf{G} = \sum_T \nabla_Y^T X$ directly on the SRAM to minimize latency, effectively implementing a batched GEMM operation, where each slice of $\mathbf{G}$ is per-sample gradient. After computation, the per-sample gradients are written back to the HBM for further processing. 

\textbf{Stage 2: Gradient Norm Computation.} The computed per-sample gradients $\mathbf{G}$ are again loaded into SRAM in smaller blocks. The norm of per-sample gradient is then computed on-chip, $\|\mathbf{G}\| = \sqrt{\sum_D \sum_P {\mathbf{G}}} \in \mathbb{R}^{B}$. 
Then, this norm calculation is stored in HBM.

\textbf{Stage 3: Gradient Clipping.} This stage involves loading both the per-sample gradients $\mathbf{G}$ and its norm $\|\mathbf{G}\|$ from the HBM into SRAM. The clipping operation is performed by computing $\mathbf{G}' = \mathbf{G} / \max\left(1, \frac{\|\mathbf{G}\|}{C}\right)$ (this division occurs in dimension B), ensuring that each gradient’s norm does not exceed the clipping threshold $C$. The clipped gradients $\mathbf{G}'$ are then stored back in HBM.

\textbf{Stage 4: Noise Addition and Aggregation.} In the final stage, the clipped per-sample gradients $\mathbf{G}'$ are loaded into SRAM, and Gaussian noise $\mathcal{N}(0, \sigma^2 C^2 \mathbf{I})$ is added to each, according to the specified noise scale $\sigma$. This process ensures differential privacy by obfuscating the contributions of individual training examples. The noisy, aggregated gradient for the weight update, $\nabla_W = \sum_B \mathbf{G}' + \mathcal{N}(0, \sigma^2 C^2 \mathbf{I})$, is computed and then written to HBM, ready for updating the model parameters.

\noindent \textbf{Limitations.}
Standard DP-SGD requires the explicit storage of per-sample gradients in HBM, which is crucial for computing the gradient norms needed for clipping. This requirement substantially increases the memory footprint. 
This method becomes impractical for LLMs, which have large model parameters and gradients due to extended sequence lengths. The extensive memory needed to store these gradients often exceeds the available HBM capacity, leading to frequent data swapping between memory and processing units, which severely slows down the training process. Crucially, the computation of gradient norms breaks down standard kernel fusion strategies, preventing the efficient integration of gradient computation and subsequent processing steps into a single operation, resulting in increased latency and inefficient GPU utilization.

\subsection{Implicit DP-SGD Workflow}
\label{sec:implicit-workflow}

Figure~\ref{fig:compare} (c) illustrates the implicit method (e.g., GhostClip, BK), which optimizes the DP-SGD process by recalculating gradients in a fused manner, thereby avoiding the explicit storage of per-sample gradients. This approach reduces memory demands but introduces computational redundancy due to multiple gradient recalculations. The implicit DP-SGD workflow is normally organized into two distinct stages:

\textbf{Stage 1: Fused Computation (corresponds to Stage 1-3 of the explicit method).} In the implicit method, stages 1 through 3 of the explicit method are executed in a fused computational process. This involves loading the activation tensor \(X \in \mathbb{R}^{B \times T \times P}\) and the output gradient tensor \(\nabla_Y \in \mathbb{R}^{B \times T \times D}\) into SRAM. The per-sample gradients tensor \(\mathbf{G} \in \mathbb{R}^{B \times D \times P}\) is recalculated by integrating gradient computation, norm calculation, and clipping into a single pass. This minimizes latency and avoids repeated data transfers to HBM. During this fused operation, the per-sample gradient norms are calculated \(\|\mathbf{G}\|\) directly on the chip. Clipping is simultaneously performed by scaling the gradients: \(\mathbf{G}' = \mathbf{G} / \max\left(1, \frac{\|\mathbf{G}\|}{C}\right)\), where \(C\) is the clipping threshold. These operations are performed without storing the intermediate states, reducing the memory footprint. 

\textbf{Stage 2: Noise Addition and Aggregation (corresponds to stage 4 of the explicit method).} The clipped gradients \(\mathbf{G}'\) are recalculated and loaded into SRAM where Gaussian noise \(\mathcal{N}(0, \sigma^2 C^2 \mathbf{I})\) is added, adhering to the specified noise scale \(\sigma\). The final aggregate gradient is then computed and written back to HBM for the model update.

\noindent \textbf{Limitations of Implicit methods:}
Implicit methods attempt to mitigate the high memory usage by segmenting the gradient computation and clipping it into several smaller, manageable tasks. However, these methods involve multiple recalculations of the per-sample gradients, which is computationally expensive.

\section{Additional Experiments Settings and Explanations}
\label{sec:exp-settings}

\subsection{Experiments Settings}

Our evaluations mainly focus on memory usage (MB) and throughput (tokens/sec) to determine the efficiency. We also show the loss of the validation data to measure the utility of private pre-training. 
Unless specified otherwise, the settings for each experiment use GPT-2 models with a sequence length of 1024, and Llama models with a sequence length of 2048, employing the AdamW optimizer as the base.

\paragraph{Batch Size \& Micro Batch Size}
For the batch size experiment, we vary the batch sizes at 1, 2, 4, and 8, using GPT-2 models of small, medium, and large scales to test the method's scalability and efficiency. Similarly, in the micro-batch size experiment, we set the micro-batch sizes at 1, 2, 4, and 8, with a gradient accumulation step of 4.

\paragraph{Experiments on Testing Utility}
We conduct an experiment to evaluate the performance of the GPT2-small model trained from scratch using DP-SGD and FlashDP under differential privacy constraints, with epsilon values set at 0.2, 0.5, and 0.8. The model is trained on the Fineweb-edu \citep{lozhkov2024fineweb-edu} dataset. Key hyperparameters include a total batch size of 524,288 tokens, a micro batch size per device of 32, and a sequence length of 1024. We use a maximum learning rate of \(6 \times 10^{-4}\) and a minimum learning rate of \(6 \times 10^{-5}\), with weight decay set at 0.1 and gradient clipping at 1.0. The model undergoes training with a validation frequency every 250 steps and model saving every 5000 steps, using both DP-SGD and FlashDP, enabling differential privacy with delta set at \(1 \times 10^{-5}\) and a clipping threshold of 100. The training aims to compare utility across different privacy levels and analyze the trade-offs between privacy and utility. We use the validation loss as the evaluation metric in Table \ref{tab:exp-precision}.

\paragraph{Distributed Training}
DDP involves distributing the model's parameters across several devices, and each device computes gradients for a subset of the data independently. This method is beneficial for managing models that fit within the memory limits of a single GPU but need faster processing through parallel execution. On the other hand, Pipeline Parallel (PP) splits the model’s layers across different devices, allowing different parts of the model to be processed simultaneously. PP is particularly useful for very large models that exceed the memory capacity of individual GPUs, enabling concurrent processing of different stages of the model across the pipeline. The experiments with DDP and PP are designed to evaluate the effectiveness of FlashDP in a distributed training context, assessing its performance in terms of memory usage and throughput across various model sizes and batch sizes. These experiments are critical to demonstrate that FlashDP can maintain its efficiency and scalability when applied to state-of-the-art LLMs, which require substantial computational resources and sophisticated training mechanisms to manage their size and complexity.

In this setup, we explore the scaling capabilities of FlashDP using DDP on four A100 GPUs (80GB each) by training GPT-2 models of small, medium, and large sizes with fixed sequence lengths of 1024 and varying batch sizes of 8, 4, and 2. Additionally, PP experiments are conducted on Llama models of sizes 3B, 7B, and 13B to evaluate throughput and memory efficiency across different stages of the model pipeline.
It is important to note that GhostClip and BK do not support the distributed modes we used.

\subsection{Additional Explanations}

GhostClip initially supports only global clipping; however, it can be easily adapted to per-layer clipping as outlined in Algorithm \ref{alg:per-layer-ghostclip}.

\begin{algorithm}[h]
\renewcommand{\algorithmicrequire}{\textbf{Require:}} 
\caption{Per-Layer GhostClip} 
\label{alg:per-layer-ghostclip} 
\begin{algorithmic}[1] 
\REQUIRE $\mathcal{L}(\theta^{(l)}, x_i)$: Loss function for layer parameters \(\theta^{(l)}\) and input \(x_i\) 
\REQUIRE $C^{(l)}$: Clipping threshold for layer \(l\) 
\REQUIRE $\sigma^{(l)}$: Noise scale for layer \(l\) 
\REQUIRE $L$: Total number of layers 
\FOR{each layer $l = 1$ to $L$} 
    \FOR{$i = 1$ to $B$} 
        \STATE Compute gradient norm for layer \(l\): \( \|g^{(l)}_i\| = \|\nabla \mathcal{L}(\theta^{(l)}, x_i)\| \) by first computing per-sample gradient in-place then computing per-sample norm.
        \STATE Clip gradient for layer \(l\): \( g'^{(l)}_i = g^{(l)}_i \min\left(1, \frac{C^{(l)}}{\|g^{(l)}_i\|_2}\right) \) by re-computing per-sample gradient $g^{(l)}_i$ in-place.
    \ENDFOR 
    \STATE Aggregate clipped gradients for layer \(l\) and add Gaussian noise: \( \tilde{g}^{(l)} = \frac{1}{B} \sum_{i=1}^B g'^{(l)}_i + \mathcal{N}(0, (\sigma^{(l)} C^{(l)})^2 I) \) 
\ENDFOR 
\end{algorithmic} 
\end{algorithm}

\section{More Experimental Results}

\subsection{Results of Micro Batch Size}
\label{sec:exp-micro-batch-size}

\begin{table*}[ht]
\centering
\caption{\textbf{Micro Batch Size Analysis.} Comparing memory and throughput at varying micro batch sizes B (1, 2, 4, 8) and the same gradient accumulation steps (4) for GPT-2 sizes with differential privacy methods under consistent settings with Table \ref{tab:result-batchsize}.}
\resizebox{\textwidth}{!}{
\begin{tabular}{ccccccc|ccccc}
\hline
\multirow{2}{*}{Model} & \multirow{2}{*}{B} & \multicolumn{5}{c}{Memory Usage (MB x1e4)} & \multicolumn{5}{c}{Throughput (tokens/sec x1e4)} \\ \cline{3-12} 
 &  & NonDP  & Opacus  & GhostClip  & BK  & FlashDP  & NonDP  & Opacus  & GhostClip  & BK  & FlashDP  \\ \hline
GPT2-small & 1 & 0.51 & 0.97(x1.90) & 0.51(x1.00) & 0.71(x1.39) & {\bf 0.51(x1.00)} & 3.07 & 1.20(x0.39) & 0.60(x0.20) & 1.75(x0.57) & {\bf 1.86(x0.61)} \\
GPT2-medium & 1 & 1.26 & 1.69(x1.34) & {\bf 1.25(x0.99)} & 1.81(x1.44) & 1.26(x1.00) & 1.27 & 0.61(x0.48) & 0.45(x0.35) & 0.86(x0.68) & {\bf 0.91(x0.72)} \\
GPT2-large & 1 & 2.48 & 3.64(x1.47) & {\bf 2.46(x0.99)} & 3.21(x1.29) & 2.48(x1.00) & 0.67 & 0.39(x0.43) & 0.32(x0.46) & 0.47(x0.69) & {\bf 0.53(x0.89)} \\ \hline
GPT2-small & 2 & 0.87 & 1.15(x1.32) & 1.00(x1.15) & 1.06(x1.22) & {\bf 0.87(x1.00)} & 3.22 & 1.68(x0.52) & 0.92(x0.29) & 1.91(x0.59) & {\bf 2.32(x0.72)} \\
GPT2-medium & 2 & 2.07 & 2.88(x1.39) & {\bf 2.01(x0.97)} & 2.62(x1.27) & 2.07(x1.00) & 1.38 & 0.88(x0.64) & 0.65(x0.47) & 0.88(x0.64) & {\bf 1.04(x0.75)} \\
GPT2-large & 2 & 3.91 & 6.07(x1.55) & {\bf 3.83(x0.98)} & 4.43(x1.13) & 3.91(x1.00) & 0.74 & 0.46(x0.62) & 0.43(0.58) & 0.49(x0.66) & {\bf 0.59(x0.80)} \\ \hline
GPT2-small & 4 & 1.53 & 2.10(x1.37) & {\bf 1.48(x0.97)} & 1.73(x1.13) & 1.53(x1.00) & 3.72 & 2.49(x0.67) & 1.50(x0.40) & 2.30(x0.62) & {\bf 2.59(x0.70)} \\
GPT2-medium & 4 & 3.58 & 5.51(x1.54) & {\bf 3.46(x0.97)} & 4.04(x1.13) & 3.58(x1.00) & 1.48 & 0.97(x0.66) & 0.86(x0.58) & 0.99(x0.67) & {\bf 1.29(x0.87)} \\
GPT2-large & 4 & 6.60 & - & {\bf 6.45(x0.98)} & - & 6.60(x1.00) & 0.79 & - & 0.53(x0.67) & - & {\bf 0.65(x0.82)} \\ \hline
GPT2-small & 8 & 2.86 & 4.00(x1.40) & {\bf 2.78(x0.97)} & 3.06(x1.07) & 2.86(x1.00) & 3.87 & 2.60(x0.67) & 1.99(x0.51) & 2.44(x0.63) & {\bf 2.73(x0.71)} \\
GPT2-medium & 8 & 6.60 & - & {\bf 6.37(x0.97)} & 7.16(x1.08) & 6.60(x1.00) & 1.55 & - & 1.03(x0.66) & 1.05(x0.68) & {\bf 1.19(x0.77)} \\
GPT2-large & 8 & - & - & - & - & - & - & - & - &- &- \\ \hline
\end{tabular}
}
\label{tab:result-microbatchsize}
\end{table*}

Table \ref{tab:result-microbatchsize} further explores the impact of varying micro batch sizes, a crucial factor for managing memory in constrained environments and optimizing the use of gradient accumulation steps. FlashDP consistently displayed minimal memory footprint increases and maintained high throughput efficiency, even as micro batch sizes increased. For example, at a micro batch size of 8 for the GPT-2 medium model, FlashDP's memory usage was $6.49 \times 10^4$ MB--marginally higher than its usage at smaller micro batch sizes and significantly lower than Opacus at the same size. This robust performance underscores FlashDP's effective management of memory, which is essential for scaling up the training of large models without excessive hardware requirements. 

To be specific, 1) Opacus showed a consistent increase in memory usage as micro batch sizes increased, which is indicative of its inefficient memory handling under fragmented gradient computations. 2) GhostClip, while better in memory usage compared to Opacus, didn't scale as well in throughput, which decreased noticeably with larger micro batches, reflecting the computational cost of gradient recalculations. 3) BK displayed trends similar to Opacus but generally used slightly less memory and provided slightly better throughput, suggesting a more optimized handling of gradient accumulation steps. 4) FlashDP maintained minimal increases in memory usage with increasing micro batch sizes and consistently provided the highest throughput, highlighting its effective integration of operations within the computational workflow. To summarize, as the micro batch size increases, FlashDP's memory usage increases only slightly and still maintains the highest throughput, demonstrating its efficient memory management techniques.

\subsection{Results of AMP Training Scalability}
\label{sec:exp-amp}

\begin{figure*}[htbp]
    \centering
    \subfigure[Memory Usage - float16]{
        \begin{minipage}{.3\textwidth}  
            \includegraphics[width=\textwidth]{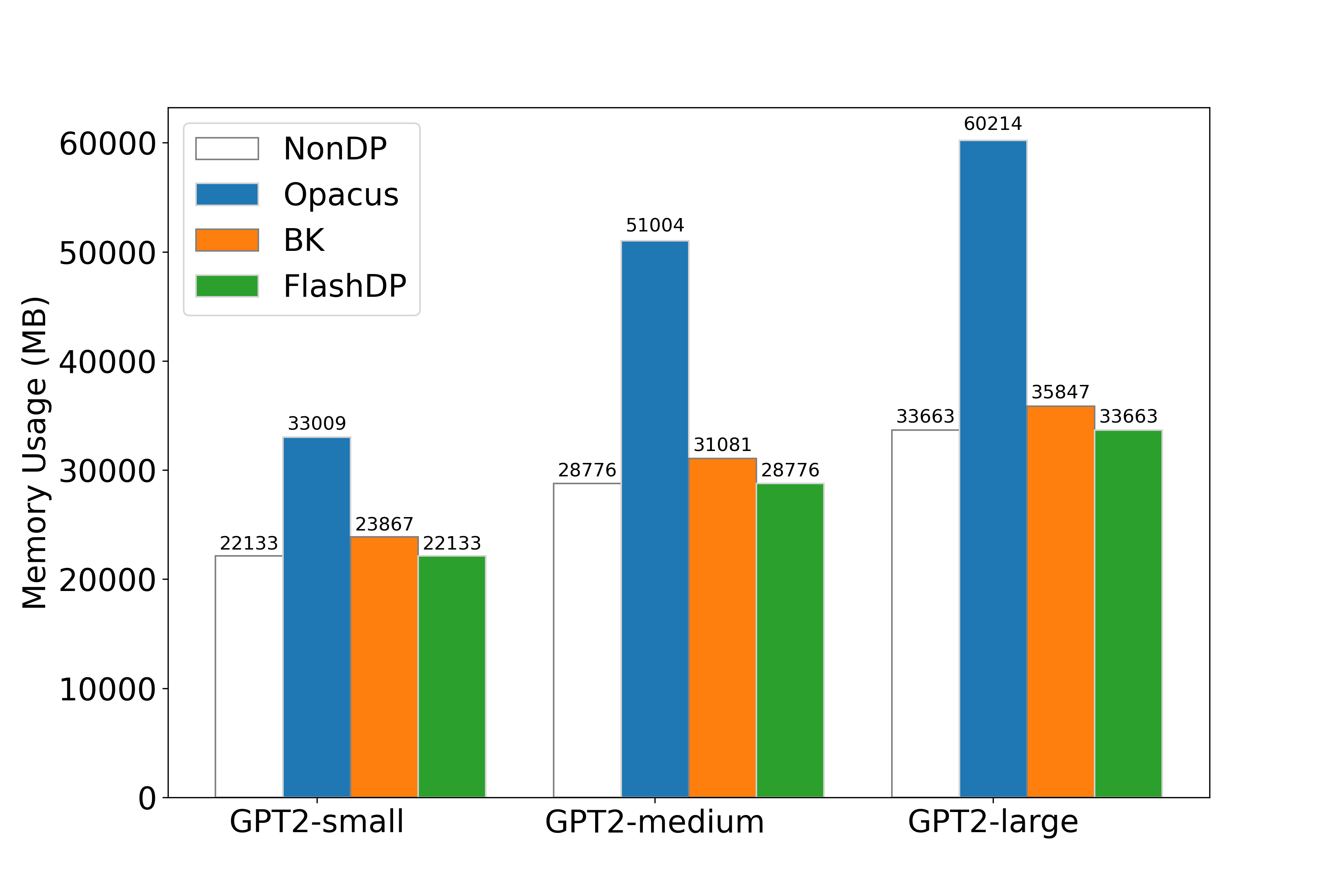}
                \label{fig:memory}
        \end{minipage}
    }
    \subfigure[Throughput - float16]{
        \begin{minipage}{.3\textwidth}
            \includegraphics[width=\textwidth]{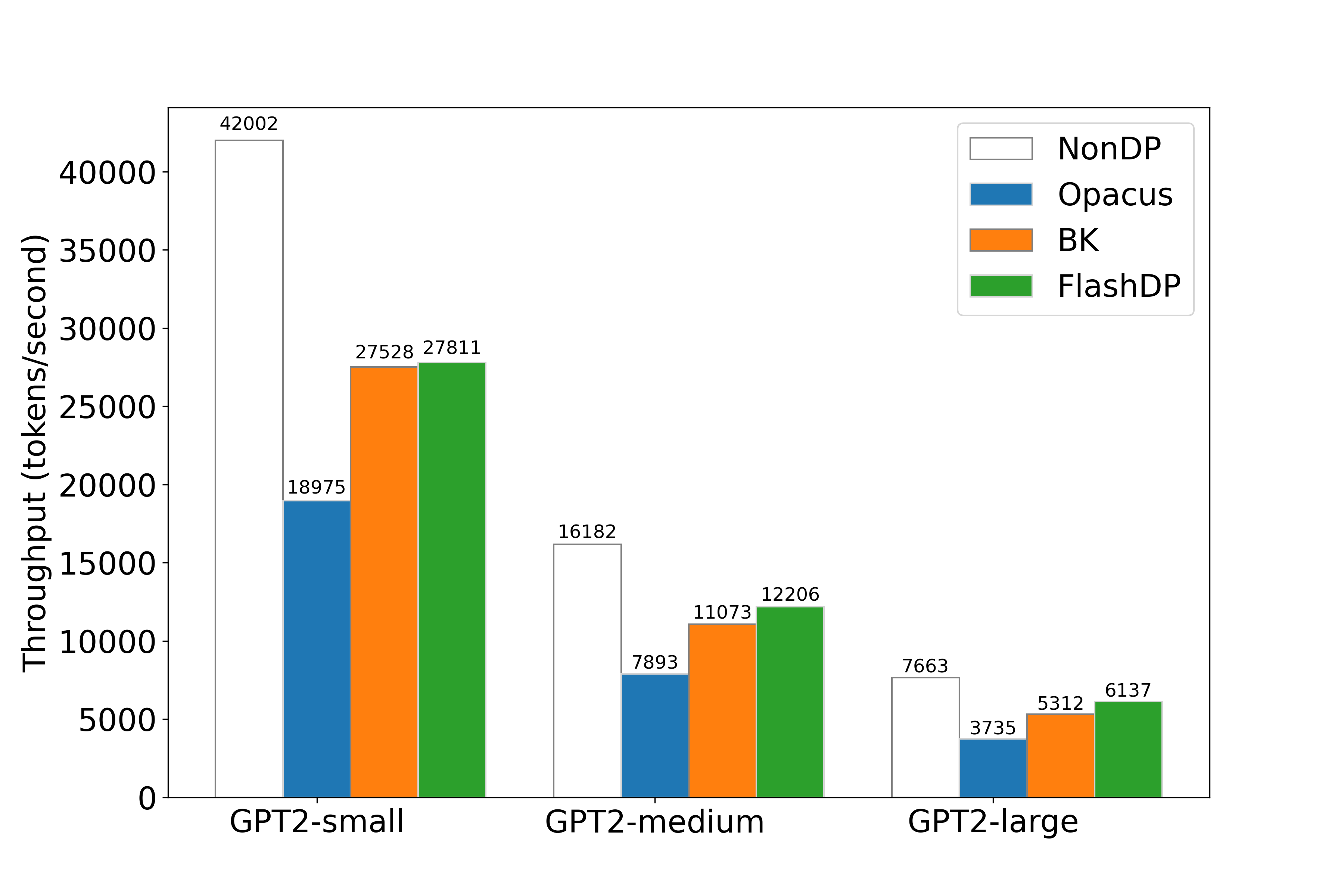}
                \label{fig:throughput}
        \end{minipage}
    }
    \subfigure[Throughput - bfloat16]{
        \begin{minipage}{.3\textwidth}
            \includegraphics[width=\textwidth]{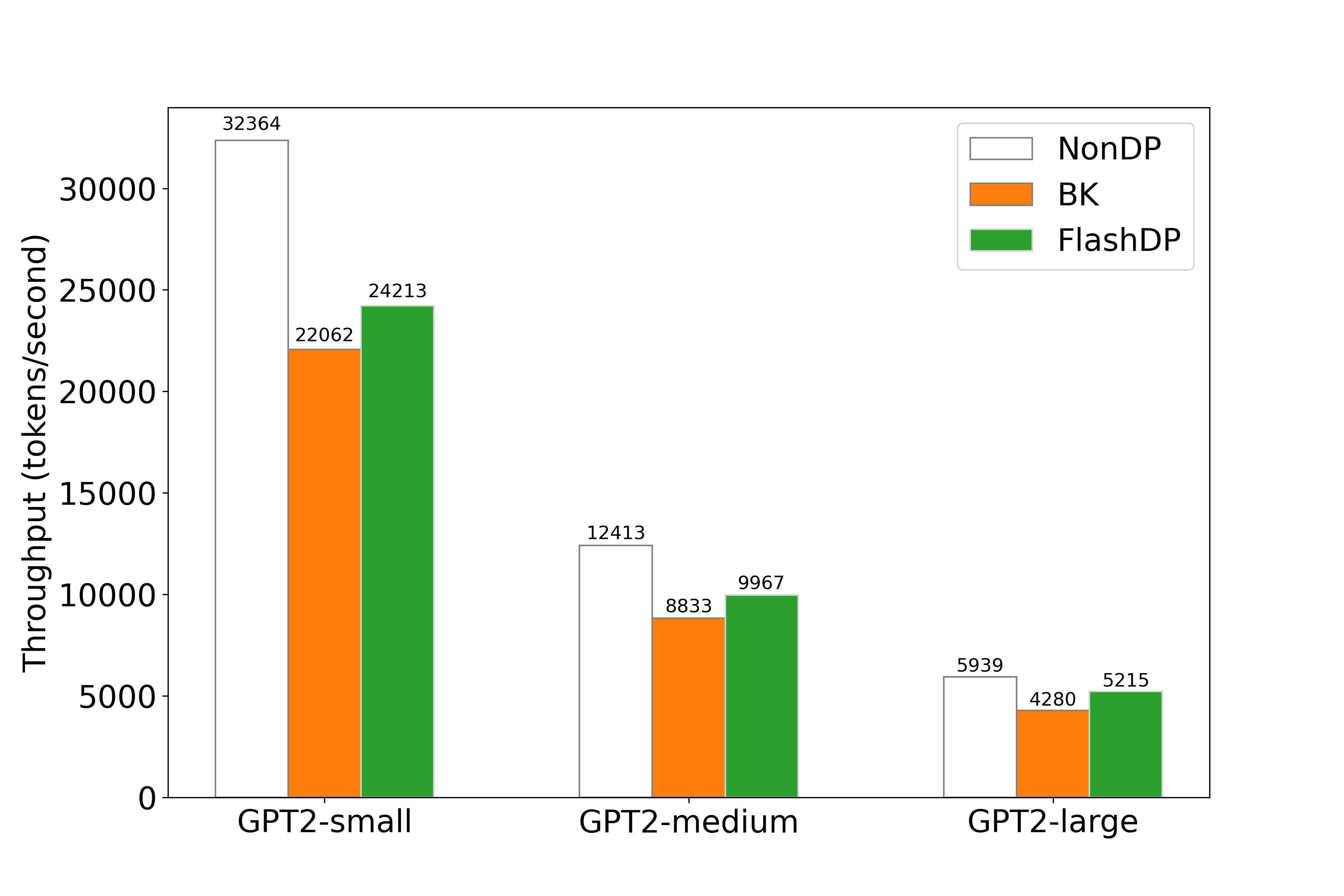}
                \label{fig:throughput}
        \end{minipage}
    }
    \caption{\textbf{Memory and Throughput Analysis of GPT-2 Models Using Automatic Mixed Precision (AMP) Training Across Float16 and BFloat16 Precision.}: (a) Demonstrates the memory usage for GPT-2 small, medium, and large models with Float16 precision. (b) shows throughput using Float16 precision, and (c) shows throughput with BFloat16 precision.}
    \label{fig:result-amp}
\end{figure*}

Automatic Mixed Precision (AMP) \citep{micikevicius2017mixed} training involves utilizing lower precision formats like float16 and bfloat16 within a training session to reduce computational demands and memory usage. This strategy is particularly valuable for large language models (LLMs), which typically require substantial computational resources. By employing AMP, training processes can be accelerated, and larger models or batches can be managed more efficiently without proportional increases in hardware capacity. The integration of differential privacy with AMP, especially in techniques like FlashDP, is critical for exploring the practical limits of DP-SGD. This experiment assesses how FlashDP adapts to AMP settings compared to other methods, and evaluates the impact on memory efficiency and processing speed, which are crucial for the scalability of private training in constrained environments.

In our experiments, we analyze GPT-2 models of varying sizes using batch sizes of 8, 4, and 2 across float16 and bfloat16 precision formats to measure memory usage and throughput, examining FlashDP's performance relative to NonDP, Opacus, and BK methods.
It is important to note that  GhostClip does not support AMP, and Opacus does not support the bfloat16 precision format.

\noindent {\bf Memory Usage Analysis.}
As depicted in Figure \ref{fig:result-amp} (a), the memory usage across GPT-2 models of different sizes indicates that FlashDP, when utilizing AMP in both float16 and bfloat16 formats, maintains lower memory consumption compared to Opacus and BK, and closely approximates the NonDP configuration. This showcases FlashDP's effective use of AMP to minimize memory overhead, facilitating the training of large models under stringent privacy constraints.

\noindent {\bf Throughput Analysis with Float16 and BFloat16.}
In terms of throughput, Figure \ref{fig:result-amp} (b) and 5(c) present a comprehensive look at the advantages of using float16 and bfloat16 precision formats under AMP. FlashDP consistently outperforms Opacus and BK in throughput metrics across both precision types. This is especially notable in larger model configurations, where the differences in throughput become more pronounced, highlighting FlashDP's capability to handle extensive computational loads efficiently. As demonstrated in Figure 5(b), FlashDP exhibits significant throughput advantages over the other DP methods. This performance is indicative of the efficient computational optimizations that FlashDP leverages within the AMP framework. As shown in Figure \ref{fig:result-amp} (c), while bfloat16 typically offers slightly lower computational throughput than float16 due to its numerical properties, FlashDP's implementation still ensures that it outperforms other differential privacy methods. This underscores FlashDP’s robust performance across varying precision settings.

\subsection{Results of Distributed Training}
\label{sec:exp-dist}

Distributed Data Parallel (DDP) \citep{li2020pytorch} and Pipeline Parallel (PP) \citep{kim2020torchgpipe} are two advanced techniques crucial for scaling the training of LLMs efficiently across multiple GPUs or nodes. 

\begin{figure*}[h]
    \centering
    \subfigure[Memory Usage]{
        \begin{minipage}{.37\textwidth}  
            \includegraphics[width=\textwidth]{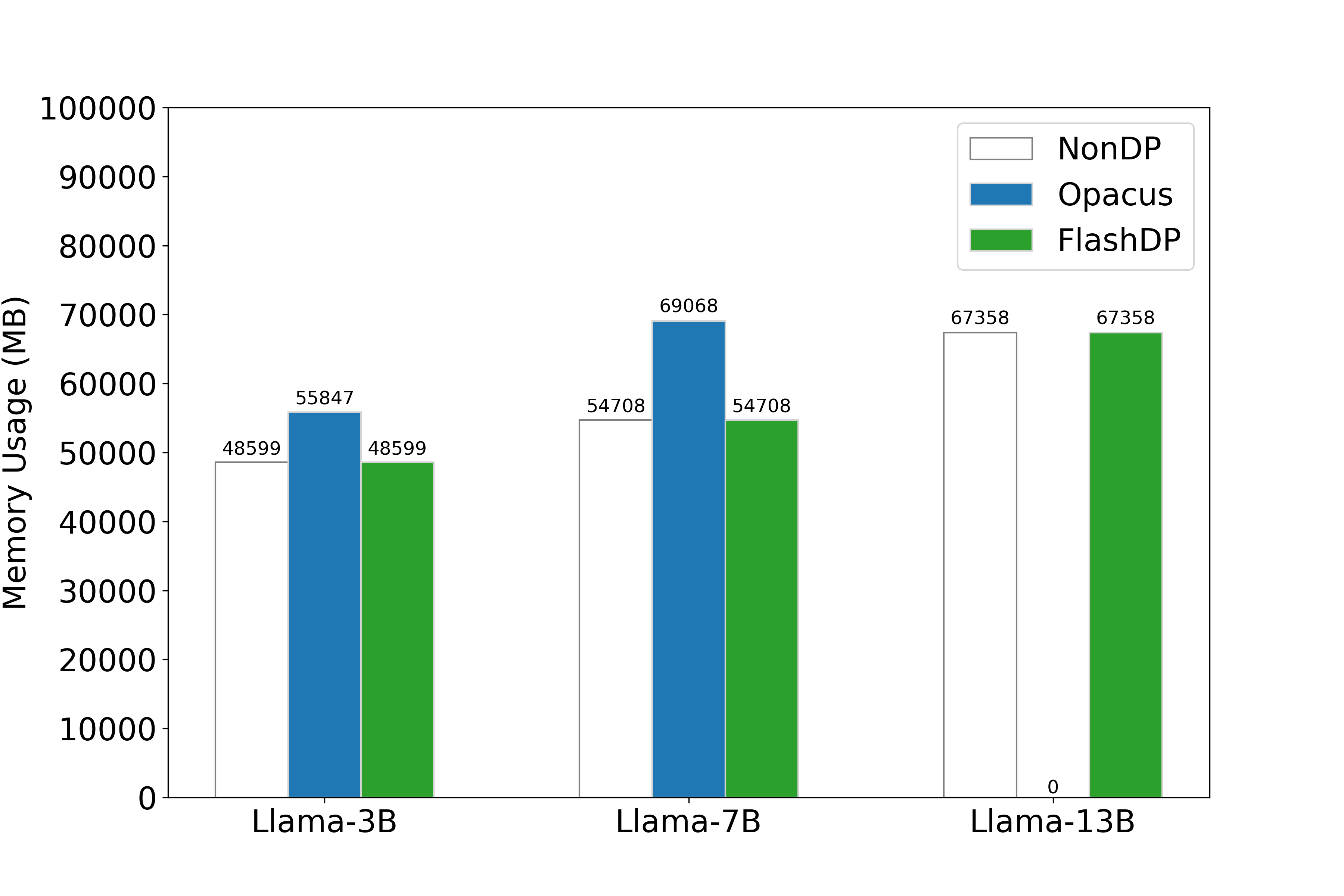}
                \label{fig:memory}
        \end{minipage}
    }
    \subfigure[Throughput]{
        \begin{minipage}{.37\textwidth}
            \includegraphics[width=\textwidth]{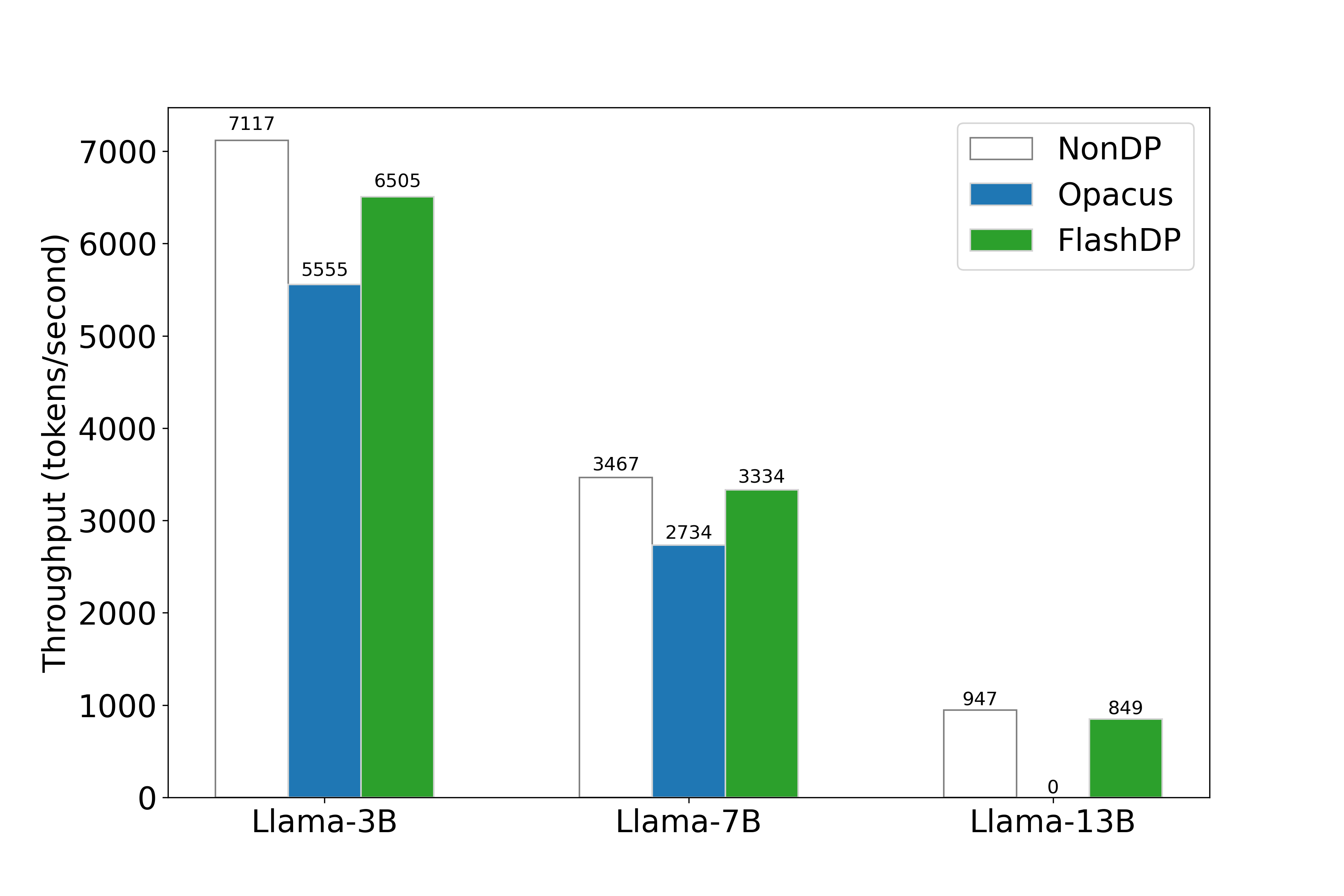}
                \label{fig:throughput}
        \end{minipage}
    }
    \caption{
    \textbf{Memory and Throughput for Llama Models Using Pipeline Parallel Training.} (a) Memory usage for Llama-3B, Llama-7B, and Llama-13B models. (b) Throughput in tokens per second across these model sizes. A value of 0 indicates out of memory.
    }
    \label{fig:result-pp}
\end{figure*}

\noindent {\bf Distributed Data Parallel (DDP).}
Figure \ref{fig:result-ddp} in Appendix illustrates the performance of different methods in a DDP setting across GPT-2 models of varying sizes. FlashDP showcases superior memory usage efficiency and higher throughput across all model sizes when compared to Opacus and BK. Notably, even as the model size increases, FlashDP maintains a competitive edge close to the NonDP benchmarks, highlighting its effective parameter distribution and gradient computation across multiple GPUs. This is crucial in scenarios where training speed and model scalability are priorities.

\noindent {\bf Pipeline Parallel (PP).}
In the PP scenario depicted in Figure \ref{fig:result-pp}, FlashDP was tested with Llama models varying from 3 billion to 13 billion parameters. The results indicate that FlashDP not only scales efficiently with increasing model size but also demonstrates significant throughput improvements compared to Opacus and BK. Particularly, FlashDP’s ability to handle the largest model (Llama-13B) with minimal throughput degradation illustrates its robustness in managing extensive computational loads, characteristic of PP environments.

\subsection{Results of Utility} \label{sec:utility}

\begin{table*}[htbp]
\centering
\caption{\textbf{FlashDP Pretrain Precision validation on GPT2-small with different privacy $\epsilon$.}}
\resizebox{.35\textwidth}{!}{
\begin{tabular}{cccc}
\hline
\multirow{2}{*}{Method} & \multicolumn{3}{c}{Validation loss} \\
 & $\epsilon=0.2$ & $\epsilon=0.5$ & $\epsilon=0.8$ \\
\hline
DP-SGD & 4.8082 & 4.8063 & 4.8061 \\
FlashDP & 4.8082 & 4.8063 & 4.8061 \\
\hline
\end{tabular}
}
\label{tab:exp-precision}
\end{table*}

In our study, FlashDP is meticulously optimized for DP-SGD, focusing on enhancing GPU I/O and system-level efficiencies without altering the fundamental algorithmic components of per-layer DP-SGD. We conducted experiments on utility with GPT-2 small to support this, whose results are shown in Table \ref{tab:exp-precision}. From the table, we can easily see that  FlashDP demonstrates an identical validation loss to that of DP-SGD across all privacy levels.

\section{Additional Tables and More Figures}

\begin{table}[htbp]
\centering
\caption{Comparison of Backward Propagation Methods.}
\label{tab:comparison}
\begin{tabular}{>{\centering\arraybackslash}m{2.0cm}ccc}
\toprule
\multirow{2}{*}{\centering Method} & \multicolumn{2}{c}{Per-sample Gradient} & \multirow{2}{*}{\centering Implicit Fusion} \\
\cmidrule(lr){2-3}
       & Cache & Recalculation & \\
\midrule
Non-DP   & \xmark & \xmark & \cmark \\
Explicit-DP & \cmark & \xmark & \xmark \\
Implicit-DP & \xmark & \cmark & \cmark \\
FlashDP  & \xmark & \xmark & \cmark \\
\bottomrule
\end{tabular}
\end{table}

\begin{figure*}[htbp]
    \centering
    \subfigure[Memory Usage]{
        \begin{minipage}{.45\textwidth}  
            \includegraphics[width=\textwidth]{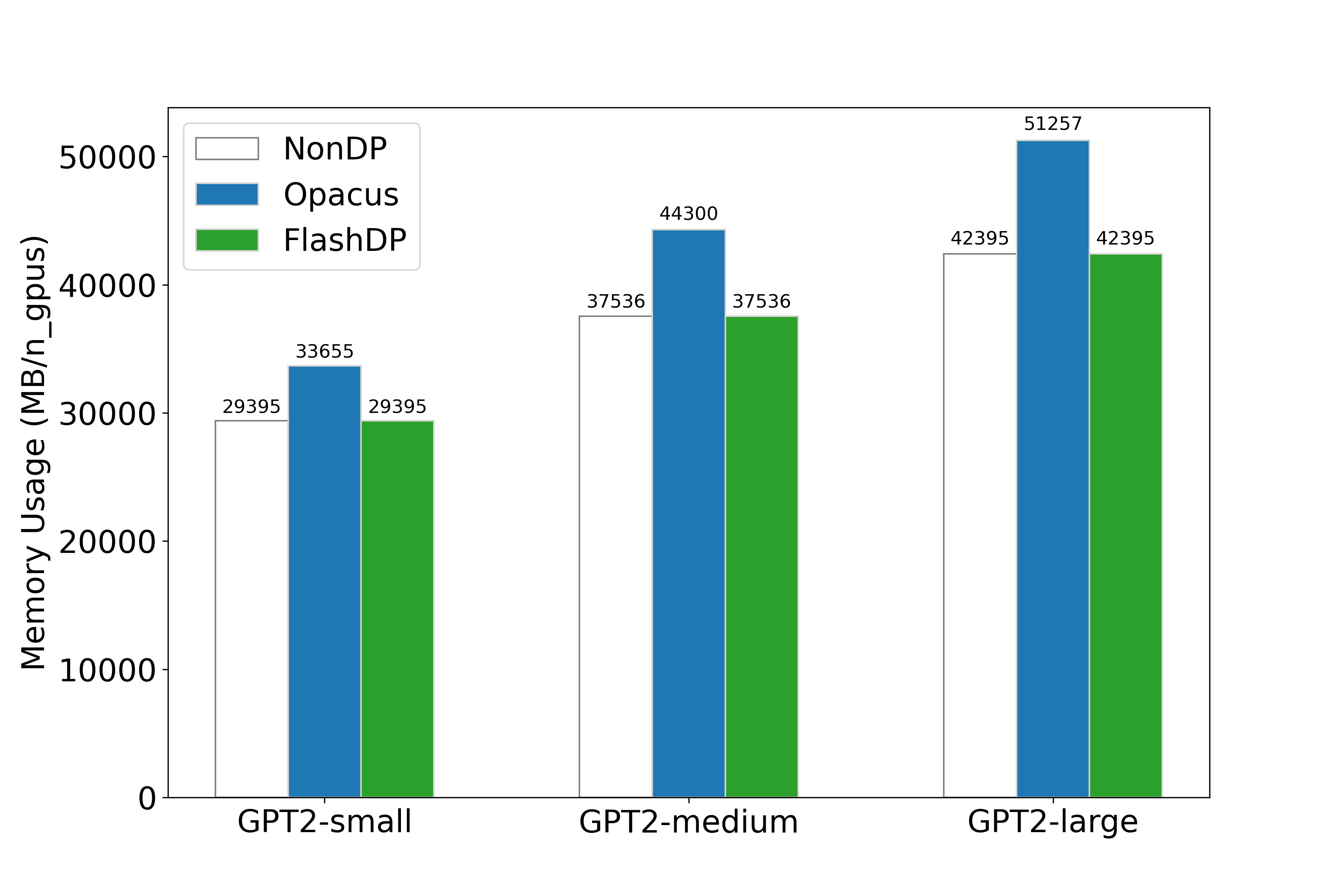}
                \label{fig:memory}
        \end{minipage}
    }
    \subfigure[Throughput]{
        \begin{minipage}{.45\textwidth}
            \includegraphics[width=\textwidth]{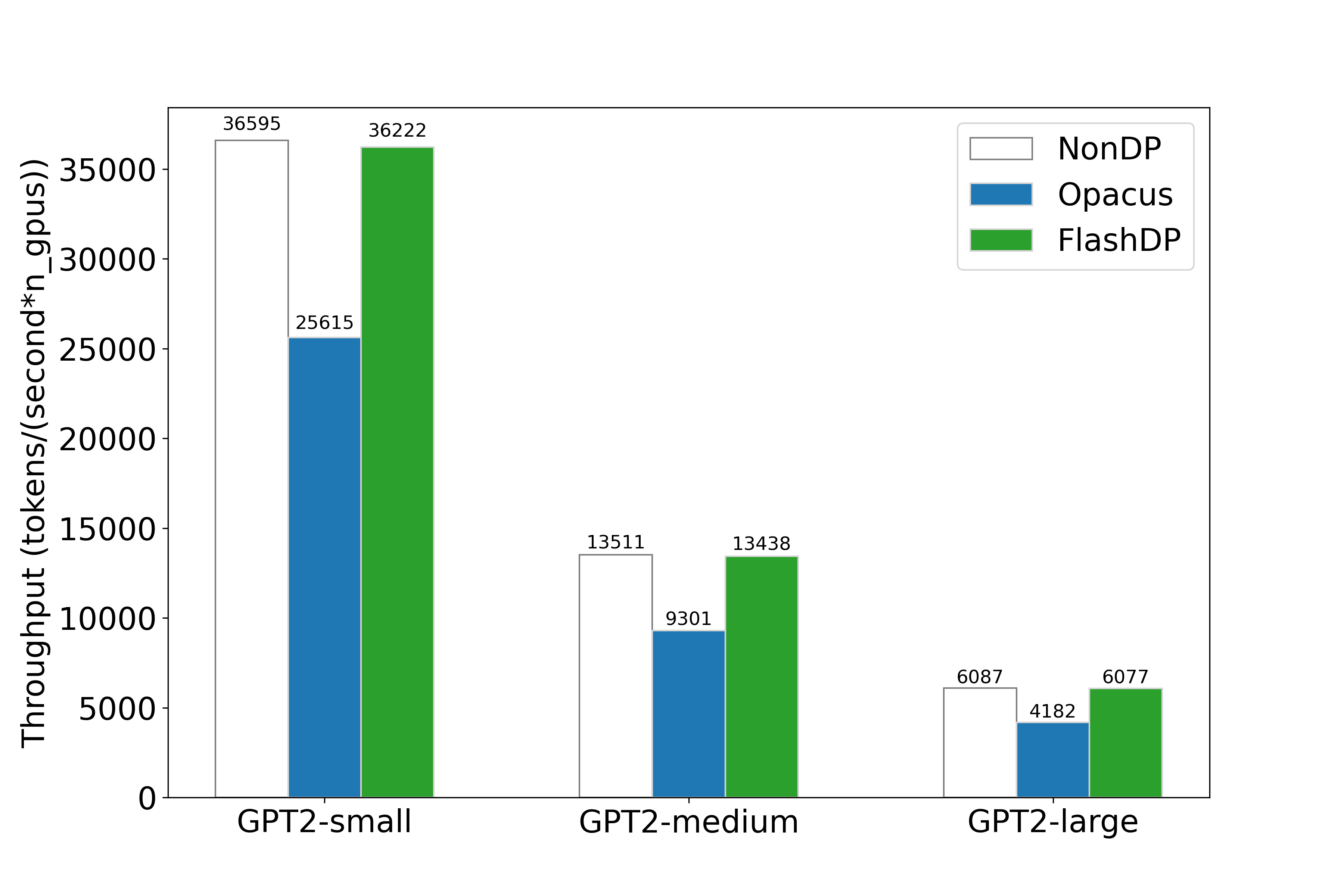}
                \label{fig:throughput}
        \end{minipage}
    }
    \caption{\textbf{Memory and Throughput for GPT Models Using Distributed Data Parallel Training.} (a) Memory usage for GPT-samll, GPT-medium, and GPT-large models. (b) Throughput in tokens per second across these model sizes. A value of 0 indicates out of memory.}
    \label{fig:result-ddp}
\end{figure*}

\end{document}